\pdfoutput=1

\documentclass[11pt]{article}

\usepackage[final]{acl}

\usepackage{times}
\usepackage{latexsym}

\usepackage[T1]{fontenc}

\usepackage[utf8]{inputenc}

\usepackage{microtype}

\usepackage{inconsolata}

\usepackage{graphicx}

\usepackage{hyperref}
\usepackage{url}
\usepackage{booktabs}       
\usepackage{amsfonts}       
\usepackage{nicefrac}       
\usepackage{xcolor}         
\usepackage{xspace}
\usepackage{enumitem}
\usepackage{tabularx}
\usepackage{booktabs}
\usepackage{multirow}
\usepackage{multicol}
\usepackage{makecell}
\usepackage{adjustbox}
\usepackage{amsmath}
\usepackage{pifont}
\usepackage{threeparttable}
\usepackage{subcaption}
\usepackage{wrapfig}
\usepackage[defaultcolor=magenta]{changes}
\usepackage{adjustbox}
\usepackage{titletoc}
\usepackage[linesnumbered,ruled,vlined]{algorithm2e}
\usepackage{amssymb}

\SetAlgoNlRelativeSize{0}          
\SetNlSty{textbf}{}{}              
\DontPrintSemicolon                
\SetInd{0.5em}{0.5em}              
\SetKwComment{tcp}{$\triangleright$\ }{}   
\SetAlgoNoEnd                      

\usepackage{float}
\usepackage[table]{xcolor}
\usepackage{makecell}
\usepackage{soul} 

\usepackage{cuted}
\usepackage{caption}
\usepackage[most]{tcolorbox}
\tcbuselibrary{listings, skins, breakable}

\tcbset{
  promptbox/.style={
    enhanced,
    colback=gray!5,
    colframe=black!60,
    listing only,
    breakable,
    pad at break=0mm,        
    listing options={
      basicstyle=\ttfamily\footnotesize,
      breaklines=true
    }
  }
}

\usepackage{diagbox}

\newcommand{\dataname}{\mbox{GroupChat-Bench}\xspace}
\newcommand{\memname}{\mbox{GroupChat-Mem}\xspace}
\title{GroupChat-Bench: Benchmarking LLM Agents on \\ Long-Term Understanding of Realistic Group Chats
}

\author{%
  Jiho Kim$^{1}$\thanks{These authors contributed equally to this work.},
  Jeewon Yang$^{1}$\footnotemark[1], Woosog Chay$^2$, Hyeonji Hwang$^1$, Daeun Kyung$^1$, \\ \textbf{Hyunseung Chung$^1$,  Eunbyeol Cho$^1$, Yeonsu Kwon$^1$, Yohan Jo$^2$, Edward Choi$^1$}\\
  $^1$KAIST $^2$SNU\\
  \texttt{\{jiho.kim, jwy24, edwardchoi\}@kaist.ac.kr} \\
}

\begin{document}
\maketitle
\begin{abstract}
  Large language model (LLM) agents are increasingly expected to participate in persistent group communication, yet existing dialogue benchmarks largely focus on short, dyadic, or single-session interactions. We introduce GroupChat-Bench, a benchmark for evaluating long-term understanding in scripted multi-party dialogues that serve as a proxy for group chats. Built from TV show scripts to avoid privacy concerns, each conversation stream spans five consecutive seasons and approximately 1,300 sessions. The benchmark tests whether agents can track participant-specific information, interpersonal relationships, and contextual changes over time. Its question pools combine refined fan-quiz-style questions with template-based questions generated from temporal knowledge graphs. Agents are evaluated as dialogue sessions unfold chronologically, with access only to the conversation history available when each question is posed. To probe the influence of memorized knowledge, GroupChat-Bench includes unanswerable questions and settings with new or swapped character names. We also propose GroupChat-Mem, which combines structured memory with a raw session repository. Experiments reveal limitations in existing memory systems, while GroupChat-Mem provides a stronger baseline. Code and data are available at \url{https://github.com/jwyang21/groupchat_bench/}.\footnote{An earlier version of this work was released as \mbox{DialSim}~\cite{kim2024dialsim}.}
\end{abstract}

\section{Introduction}
\begin{figure*}[t]
    \begin{center}
    \includegraphics[width=1.0\linewidth]{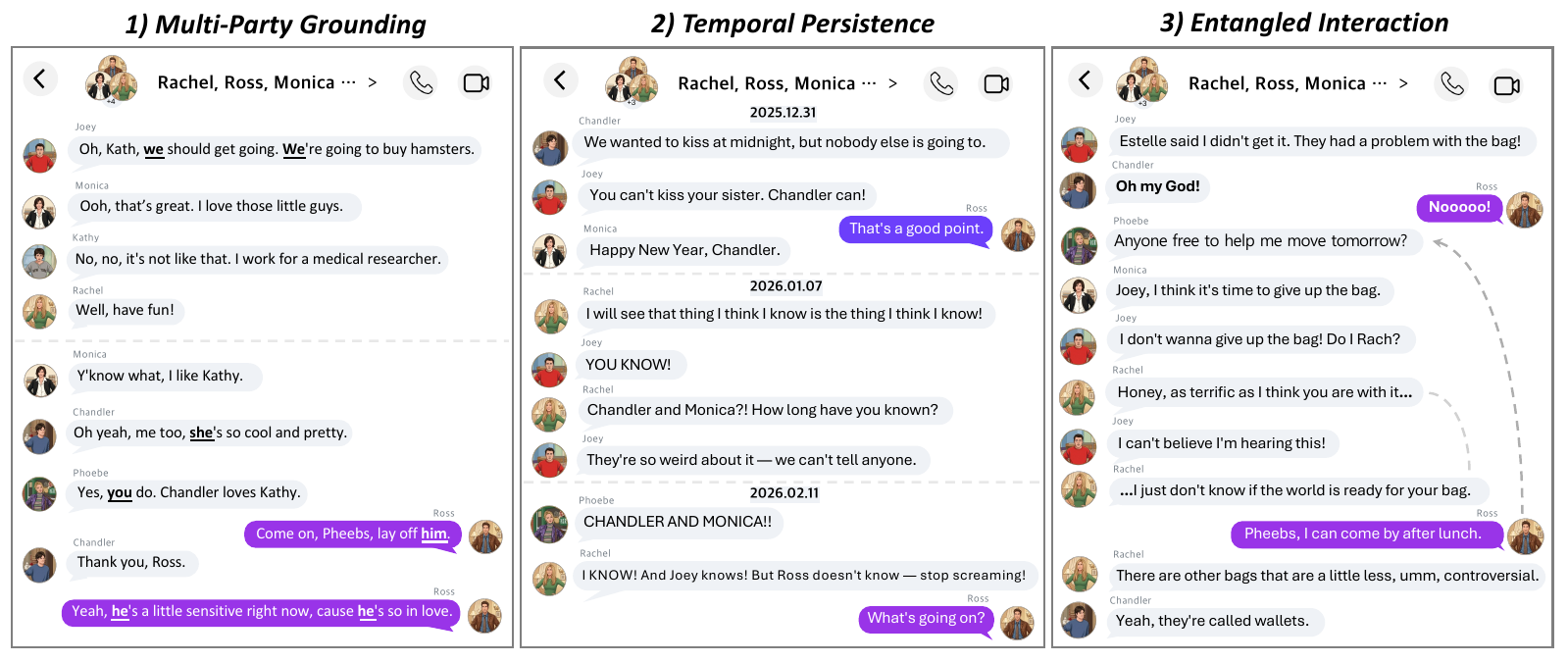}
    \end{center}
    \vspace{-2mm}
    \caption{Key challenges of group chats. (1) \textit{Multi-party grounding}: pronouns and address terms (underlined) must be resolved across speakers. (2) \textit{Temporal persistence}: facts emerge, spread, and remain unevenly known across sessions, as the Chandler--Monica relationship is discovered by Rachel and later Phoebe, while Ross still remains unaware. (3) \textit{Entangled interaction}: pile-on reactions (bold), interleaved topics (dashed arrow), and fragmented messages (dashed line) disrupt clean turn-taking, requiring implicit context to be recovered from noisy exchanges.}
    \label{figure1}
\end{figure*}

Large language model (LLM) agents are increasingly deployed in shared digital environments, including household assistants used by multiple family members and enterprise agents integrated into collaboration platforms~\cite{panay2025alexaplus, slack2026aiagents, microsoft2026copilotteams, zoom2026aicompanion}. In such settings, communication unfolds among multiple participants over time, requiring agents to move beyond one-to-one interaction models and engage in sustained group communication.

Such group interactions exhibit three characteristics that pose key challenges for LLM agents, as illustrated in Figure~\ref{figure1}. First, they require \textbf{multi-party grounding}: agents must determine not only what each utterance means, but also who it concerns, how it is addressed, and how it fits within the relationships among participants~\cite{le-etal-2019-speaking, qamar-etal-2023-speaking}. Second, group chats exhibit \textbf{temporal persistence}: personal and relational information accumulates over time, changes, and can be reinterpreted across sessions, requiring long-term contextual knowledge at both individual and group levels~\cite{10.1145/1180875.1180944, 10.1145/3274465}. Third, group conversations are highly \textbf{entangled}: fragmented utterances, pile-on reactions, and interleaved topical threads disrupt clean turn-taking, requiring agents to disentangle salient context from noisy exchanges~\cite{elsner-charniak-2008-talking, mehri-carenini-2017-chat}.

However, existing dialogue understanding benchmarks do not sufficiently reflect these properties. Many focus on short, single-session, or two-party conversations~\cite{li2017dailydialog, zhang2018personalizing, sun-etal-2019-dream, cui-etal-2020-mutual}; longer-context benchmarks often remain two-party or rely on overly clean synthetic dialogues~\cite{jang-etal-2023-conversation, maharana2024evaluating, wulongmemeval}; and existing multi-party datasets are organized as independent scenes or short episodes, leaving  memory over long-term group histories untested~\cite{yang-choi-2019-friendsqa, poria-etal-2019-meld, li-etal-2020-molweni}.

To address this gap, we propose GroupChat-Bench, a benchmark for evaluating long-term understanding of scripted multi-party dialogue as a \textit{proxy} for group chats. We use TV show scripts to provide human-authored, long-horizon interactions among recurring participants without the privacy concerns of real chat logs. While these dialogues exhibit several challenges relevant to group communication, their generalizability to naturally occurring group chats remains untested. All non-verbal descriptions (\textit{e.g.}, gestures, scene notes) are removed, and every question is verified to be answerable from the utterance text alone (\S\ref{datacon}). Each resulting conversation stream spans five consecutive seasons ($\sim$1,300 sessions on average), over which the benchmark evaluates whether agents can track participant-specific information, interpersonal relationships, and contextual changes over time.

Question pools combine refined fan-quiz questions with template-based questions from temporal knowledge graphs extracted from the dialogue. During evaluation, agents process sessions chronologically and answer questions using only the history available at query time. To mitigate data contamination, we include unanswerable questions and settings with new or swapped character names.

We further propose GroupChat-Mem, a hybrid memory system for LLM agents that combines structured memory with a repository of raw dialogue sessions to support both high-level reasoning and fine-grained evidence retrieval. Experiments on GroupChat-Bench demonstrate that existing memory systems remain limited in long-term multi-party dialogue understanding, while GroupChat-Mem provides a stronger baseline for this setting.

\section{Related Work}
\newcommand{\cmark}{\ding{51}}
\newcommand{\xmark}{\textcolor{gray!60}{\ding{55}}}
\newcommand{\namark}{\textcolor{gray!60}{--}}

\begin{table*}[t]
\centering
\normalsize
\resizebox{\textwidth}{!}{
\begin{tabular}{lccccc}
\toprule
\multirow{2}{*}{\textbf{Dataset}}
& \multirow{2}{*}{\textbf{Multi-Party Grounding}}
& \multicolumn{2}{c}{\textbf{Temporal Persistence}}
& \multirow{2}{*}{\textbf{Entangled Interaction$^\dagger$}}
& \multirow{2}{*}{\textbf{QA Pairs}} \\
 \cmidrule(lr){3-4}
&
& Dialogue Length
& Session Continuity
&
& \\
\midrule
\multicolumn{6}{l}{\textit{Short or session-independent dialogues}} \\
DREAM~\cite{sun-etal-2019-dream}
& \xmark & 0.09k & \namark & \xmark & \cmark \\
MELD~\cite{poria-etal-2019-meld}
& \cmark & 0.14k & \namark & \cmark & \xmark \\
Molweni~\cite{li-etal-2020-molweni}
& \cmark & 0.15k & \namark & \cmark & \cmark \\
FriendsQA~\cite{yang-choi-2019-friendsqa}
& \cmark & 0.43k & \namark & \cmark & \cmark \\
\midrule
\multicolumn{6}{l}{\textit{Long-term, but dyadic and synthetic}} \\
Conversation Chronicles~\cite{jang-etal-2023-conversation}
& \xmark & 1.0k & \cmark & \xmark & \xmark \\
LoCoMo~\cite{maharana2024evaluating}
& \xmark & 16k & \cmark & \xmark & \cmark \\
LongMemEval~\cite{wulongmemeval}
& \xmark & 115k, 1.5M & \xmark & \xmark & \cmark \\
\midrule
\textbf{GroupChat-Bench (ours)}
& \cmark & \textbf{350k} & \cmark & \cmark & \cmark \\
\bottomrule
\end{tabular}
}
\caption{
Comparison with existing dialogue understanding benchmarks across key conversational characteristics.
Dialogue Length denotes the approximate number of tokens in the evaluation context;
Session Continuity indicates whether sessions form a continuous chronological history involving the same participants (`\namark': single-session).
$^\dagger$Entangled Interaction refers to fragmented turn-taking, pile-on reactions,
and interleaved conversational topics (Figure~\ref{figure1});
synthetic LLM-generated dialogues are marked as lacking these phenomena when they are not represented.
}
\label{tab_conversation_datasets}
\end{table*}

\paragraph{Dialogue Understanding Benchmarks}
Existing dialogue understanding benchmarks mainly focus on short, single-session, or two-party conversations~\cite{lowe-etal-2015-ubuntu, li2017dailydialog, zhang2018personalizing, sun-etal-2019-dream, cui-etal-2020-mutual}.
Multi-party dialogue datasets~\cite{yang-choi-2019-friendsqa, poria-etal-2019-meld, li-etal-2020-molweni} model interactions among multiple speakers, but are organized as independent scenes or short episodes, leaving long-term accumulated group histories unevaluated.
\mbox{FriendsQA}~\cite{yang-choi-2019-friendsqa} shares our source material, yet evaluates span extraction within single scenes ($\sim$0.4k tokens) and retains original character names; we instead pose questions over chronologically accumulating histories ($\sim$350k tokens) under contamination controls (§\ref{charnametr}).
\mbox{SHARE}~\cite{kim-etal-2025-share} also uses scripted dialogue, but focuses on shared memories between two individuals across independent films for response generation rather than long-term multi-party QA.
Earlier long-term memory benchmarks~\cite{jang-etal-2023-conversation, maharana2024evaluating, wulongmemeval, hu2025evaluating} primarily study dyadic or synthetic dialogue histories, offering limited coverage of the fragmented, entangled exchanges found in human-authored multi-party dialogue (Table~\ref{tab_conversation_datasets}).
More recent benchmarks, \mbox{EverMemBench}~\cite{hu2026evermembench} and \mbox{GroupMemBench}~\cite{yang2026groupmembench}, also evaluate long-term multi-party memory using synthetic conversations. EverMemBench emphasizes collaborative recall, memory awareness, and profile understanding, while GroupMemBench focuses on group dynamics, speaker-grounded belief tracking, and audience-adapted language.
In contrast, GroupChat-Bench uses human-authored scripted dialogue and evaluates long-term understanding through questions whose answerability depends on the dialogue observed by the agent at query time.

\noindent\textbf{Agentic Memory Systems}~~
Agentic memory systems have been developed to enable LLM agents to maintain information across long-term interactions by storing, updating, and retrieving information from past contexts.
\mbox{HippoRAG 2}~\cite{gutierrez2025rag} continually indexes new information into a knowledge graph and retrieves it via multi-hop graph search based on Personalized PageRank~\cite{haveliwala2002topic}, while \mbox{Mem0}~\cite{chhikara2025mem0} extracts salient facts from conversations and consolidates them through LLM-driven add/update/delete decisions relative to existing memories.
\mbox{A-MEM}~\cite{xu2026mem} instead organizes memory as a network of interlinked notes, where new information triggers both the creation of new links and the evolution of existing notes.
\mbox{LD-Agent}~\cite{li-etal-2025-hello} maintains event and persona memories for personalized long-term dialogue, but focuses on dyadic response generation rather than tracking relationships among multiple observed participants.
We benchmark general-purpose memory systems that can be applied to long-term multi-party dialogue streams to assess their ability to handle this setting.

\section{GroupChat-Bench}
\subsection{Data Construction}
\label{datacon}
\begin{figure*}[t]
    \centering
    \begin{center}
    \includegraphics[width=\linewidth]{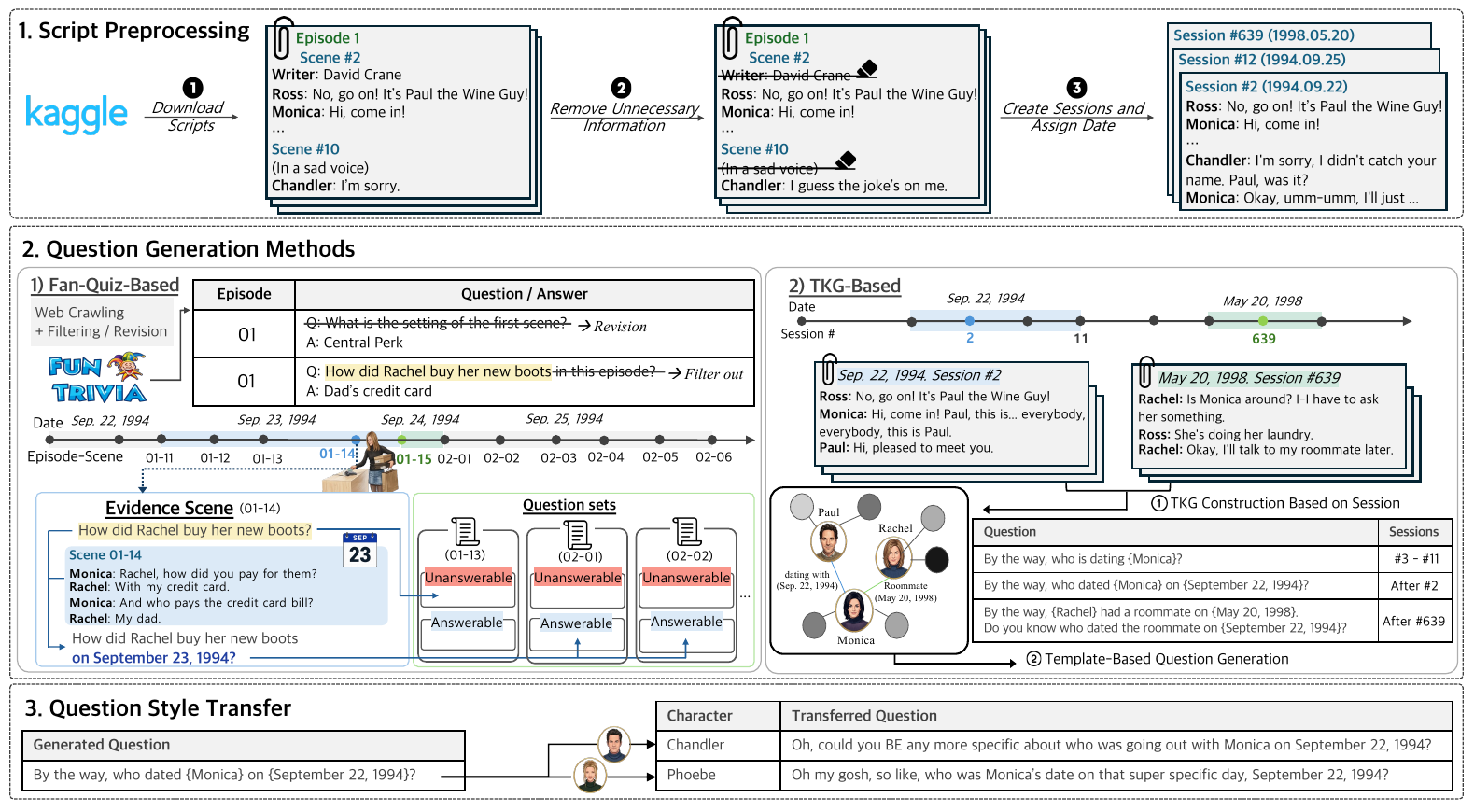}
    \end{center}
    \vspace{-4.5mm}
    \caption{Overview of the \dataname~construction pipeline, including script preprocessing, fan-quiz- and TKG-based question generation, and character style transfer. Character names are subsequently anonymized (New Names) or swapped among characters (Swapped Names) (§\ref{charnametr}); original names shown for illustration.}
    \vspace{-2.5mm}
    \label{figure2}
\end{figure*}

\dataname~was developed using scripts of three popular TV shows (\textit{i.e.}, Friends, The Big Bang Theory, and The Office\footnote{The scripts were downloaded from the website Kaggle.}). These scripts were first preprocessed to serve as dialogue data. Next, questions were generated for each script, drawing from fan quizzes and a temporal knowledge graph (TKG). We then applied character style transfer and anonymized character names using new names, with an adversarial setting based on name swapping. Figure~\ref{figure2} illustrates the construction process.

\subsubsection{Script Preprocessing}
\label{scriptprepro}
Each script spans five consecutive seasons ($\sim$20 episodes per season), with each episode composed of multiple scenes. Beyond utterances, scripts contain non-verbal descriptions of characters' actions, expressions, and scenes, as well as production metadata (\textit{e.g.}, writer names). We remove all such content to create $Script_{pre}$, which consists solely of speaker-attributed utterances, mirroring a text-based group chat in which participants observe only each other's messages; any information conveyed non-verbally in the original show is thus excluded from the dialogue stream and, by construction, from the question pool (\S\ref{fanquiz}, \S\ref{tkgq}).

We then selected scenes involving the main character (\textit{i.e.}, Friends: Ross, The Big Bang Theory: Sheldon, The Office: Michael\footnote{The characters with the most lines were selected.}) from $Script_{pre}$ and sequentially numbered them as dialogue sessions $S_i$, resulting in the final dialogue $\mathcal{D} = \{\mathcal{S}_1, \mathcal{S}_2, ..., \mathcal{S}_N\}$. From this point onward, we refer to these selected scenes as sessions. Each selected session retains all speaker-attributed utterances from the original scene, including those of other participants, rather than only the main character's utterances. This session extraction preserves a coherent dialogue flow from the main character’s perspective while also enabling the construction of unanswerable questions. Specifically, scenes without the main character are excluded from the agent’s observed dialogue stream but retained for evidence annotation, allowing us to create questions about facts that exist in the script yet remain inaccessible to the agent.
Additionally, since some of our questions involve time conditions, we manually assigned a date to each session in $\mathcal{D}$, determined from conversation content and episode air dates; the specific assignment rules are detailed in Appendix~\ref{app:daterule}.

\subsubsection{Fan Quiz-Based Question Generation}
\label{fanquiz}
We utilized a fan quiz website FunTrivia\footnote{https://www.funtrivia.com/} to generate our questions.
Fan quizzes cover a range of difficulty levels and focus on major events from each episode, making them promising for evaluating dialogue comprehension. 
We began by extracting episode-specific quizzes from the site. Since these quizzes were created by dedicated fans, many required knowledge unrelated to the dialogue itself (\textit{e.g.}, ``What is the name of the actor who played the clerk?''). To filter out these questions, we first selected quizzes that could be answered by referencing $Script_{pre}$ using an LLM. The LLM also annotated the evidence scenes in $Script_{pre}$ that support each question.\footnote{Fan quizzes exist for each episode, so we annotated them based on $Script_{pre}$ and then matched them to sessions of $\mathcal{D}$.} The authors then manually corrected the LLM annotations for quiz filtering and evidence-scene mapping. Note that since both procedures are performed against $Script_{pre}$, the retained questions rely solely on the utterance text, not on non-verbal information.

We then labeled each question as answerable or unanswerable for each session, based on whether the main character had observed all corresponding evidence scenes before the query time. If the evidence appeared only outside the main character’s observed sessions (\textit{i.e.}, $\mathcal{D}$), the question was labeled unanswerable. We also generated temporal variants by adding session dates to questions, enabling evaluation of long-term memory (\textit{e.g.,} ``How did Rachel buy her new boots on September 23, 1994?'').
Detailed question generation processes are provided in Appendix~\ref{app:fangen}.

\subsubsection{Temporal Knowledge Graph-Based Question Generation}
\label{tkgq}
Fan quizzes are useful for generating our questions, but since they are episode-specific and user-generated, the questions don't span multiple episodes and the number of questions is limited ($\sim$1k). To address this, we constructed a knowledge graph for each session and used it to generate questions.
Initially, we used the LLM to extract triples (\textit{i.e.}, [head, relation, tail]) from each session $S_i$ in $\mathcal{D}$. The authors then manually reviewed all extracted triples and removed incorrectly extracted ones. We employed 32 relations (\textit{e.g.}, girlfriend) derived from DialogRE~\citep{yu-etal-2020-dialogue}, a high-quality dataset where human annotators manually extracted relations from Friends scripts, classifying relationships between characters into 36 categories.
We adapted and modified these relations for our purpose. 
More details about the relations are provided in Appendix~\ref{app:tkgrel}.
Finally, we combined the triples from each session with their dates to create a temporal knowledge graph (TKG) composed of quadruples (\textit{i.e.}, [head, relation, tail, date]).

Using the TKG, we generated questions for each session by sampling one-hop or two-hop quadruples and checking whether the main character could answer them at the query time. Because the question form and answerability depend on when the question is asked, the same fact can be phrased differently across sessions: [Rachel, boyfriend, Ross, 1994-08-08] yields ``Who was Rachel’s boyfriend on August 8th, 1994?'' when asked later, but “Who is Rachel’s boyfriend?” when asked on that date. If no prior evidence exists, the correct answer is ``I don’t know.'' We manually verified the answers and applied the same procedure to two-hop questions. Examples of question templates and generated questions are provided in Appendix~\ref{app:tkgqexam}.

\subsubsection{Answer Choices Generation}
\label{finaldatapro}
To create multiple-choice questions, we crafted answer choices for each question. All questions include the choice ``(E) I don't know.'', which is the gold answer for unanswerable questions. For fan-quiz questions, the four choices and correct answers follow the original quiz. For TKG-based questions, distractors are sampled from tails of other quadruples sharing the same relation; to make the test more adversarial, we prioritize the subject's own past or future partners (\textit{e.g.}, Rachel's past or future boyfriends for ``Who is Rachel's boyfriend?''), so that memorized or temporally misplaced knowledge points to a wrong choice.

\begin{figure*}[t]
    \centering
    \begin{center}
    \includegraphics[width=\linewidth]{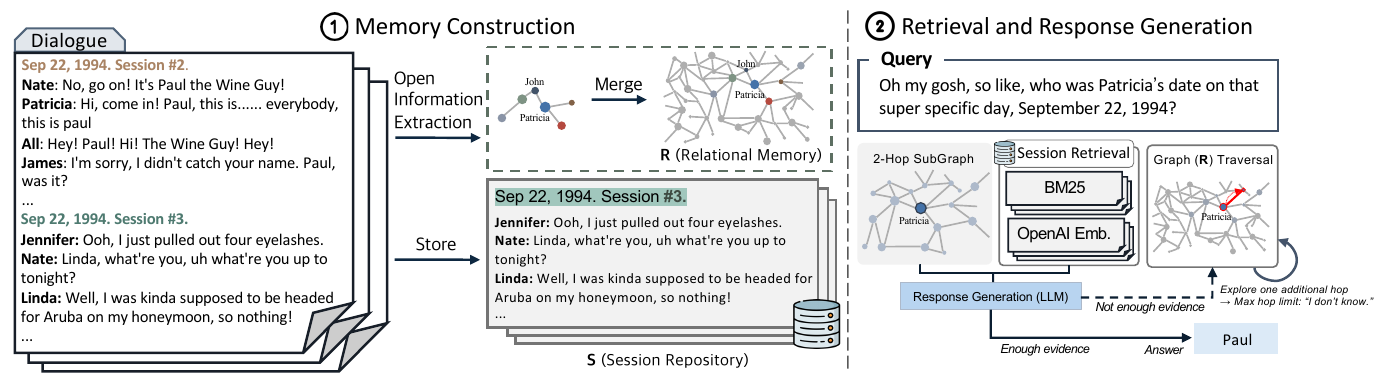}
    \end{center}
    \vspace{-4.8mm}
    \caption{Overall architecture of GroupChat-Mem.
(1)~An LLM performs Open Information Extraction (schema-free
extraction of [subject, relation, object] records) over incoming
sessions to build the relational memory~$\mathcal{R}$, while raw
sessions are stored in the session repository~$\mathcal{S}$.
(2)~Given a query, evidence is retrieved from both $\mathcal{R}$ and
$\mathcal{S}$, with additional graph traversal over $\mathcal{R}$ when
the initial evidence is insufficient.}
    \vspace{-2mm}
    \label{figure3}
\end{figure*}

\subsubsection{Question Style Transfer}
Although fan quiz-based and TKG-based questions provide broad coverage, their quiz-like or templated structure can result in artificial phrasing. To make the questions more natural, we apply character style transfer, rewriting each question in the distinctive tone of the corresponding character so that the characters themselves appear to be asking (\textit{e.g.}, Generic style: ``How did Rachel buy her new boots?''$\rightarrow$ Style of Joey Tribbiani from Friends: ``Hey, how did Rachel manage to snag those killer boots, huh?''). This transformation is powered by the LLM, and subsamples are reviewed by the authors to ensure that the original intent was preserved.
More examples of style-transferred questions for each character are in Appendix~\ref{app:styletrans}.

All LLM-assisted procedures in the data construction process were performed using GPT-4~\citep{openai2023gpt4}.

\subsubsection{Character Name Transformation}
\label{charnametr}
To reduce agents' reliance on prior knowledge of the TV shows, we replace original character names with new names unseen in the source shows (\textit{e.g.}, Joey
$\rightarrow$ John); we refer to this setting as \textbf{New Names}
and use it as the default in all experiments unless otherwise noted.
We additionally create a more adversarial \textbf{Swapped Names}
setting by exchanging names among characters (\textit{e.g.}, Joey
$\leftrightarrow$ Monica), which makes memorized character knowledge
actively misleading, as it now contradicts the dialogue context.

\subsection{Agent Evaluation}
The final dataset consists of the dialogue $\mathcal{D}$ and session-specific candidate question sets. Specifically, $\mathcal{D}$ contains $N$ dialogue sessions, and each session $S_i$ is associated with a candidate question set $\mathcal{Q}_i$. Each $\mathcal{Q}_i$ contains question-answer pairs $(q_{i,j}, a_{i,j})$, where $q_{i,j}$ is a question that can be asked during $S_i$ and $a_{i,j}$ is the corresponding gold answer.

Agents process sessions in chronological order and answer one question per session. At each session \(S_i\), we sample a question-answer pair \((q_i,a_i)\) from \(Q_i\) under a stratified sampling policy, targeting 50\% fan-quiz questions and 20\% unanswerable questions to ensure coverage of both question sources and answerability conditions. Each question is posed at a randomly sampled point within the session, and the agent can use only the dialogue history observed up to that point. We repeat this procedure three times with independently sampled question sets, using the same questions for all memory systems within each run. Reported accuracies are averaged over the three runs and characterize performance under this stratified evaluation protocol rather than uniform sampling over the full candidate pool.

\renewcommand{\arraystretch}{1.35}
\begin{table}[t]
\centering
\resizebox{\linewidth}{!}{%
\begingroup
\begin{tabular}{lccc}
\toprule
 & \textbf{FR} & \textbf{TBBT} & \textbf{TO} \\ \toprule
Dialogue Length (tokens)  & 335,439 & 367,636 & 352,914 \\ 
\# of Dialogue Sessions & 788 & 805 & 2,347 \\ \hline
Fan Quiz Questions\textsuperscript{\textdagger} & 192.9 & 26.7 & 42.7 \\ 
TKG Questions\textsuperscript{\textdagger} & 1173.2  & 1280.1  & 455.1 \\ 
Total Question Candidates\textsuperscript{\textdagger} & 1366.1 & 1306.8  & 497.9 \\ 
\enspace $\hookrightarrow$Answerable Questions\textsuperscript{\textdagger} & 1215.0 & 1239.7 & 410.9 \\
\enspace $\hookrightarrow$Unanswerable Questions\textsuperscript{\textdagger} & 151.1 & 67.2 & 86.9 \\ \hline
\textbf{Approx. \# of Possible Tests} & $\mathbf{1366.1^{788}}$ & $\mathbf{1306.8^{805}}$ & $\mathbf{497.9^{2347}}$ \\ \toprule
\end{tabular}%
\endgroup
}
\vspace{-3mm}
\caption{Statistics of \dataname. $\dagger$ denotes the average number of questions per session. FR: Friends, TBBT: The Big Bang Theory, TO: The Office.}
\vspace{-2mm}
\label{stattable}
\end{table}

\subsection{Statistics and Evaluation Formats}
\label{datastat}
Table~\ref{stattable} presents the statistics of \dataname, including the number of dialogue sessions and the average number of question candidates per session. \dataname~ supports both \textbf{multiple-choice} and \textbf{open-ended} question formats for evaluation. In the multiple-choice setting, each question provides five answer choices. During evaluation, unanswerable questions are sampled to constitute 20\% of the test questions, balancing the five answer choices rather than reflecting their natural frequency in the candidate pool. 
In the open-ended setting, we used the token-level F1/EM (Exact Match) metrics from SQuAD~\citep{rajpurkar2016squad} for evaluation.
The detailed evaluation protocol is provided in Appendix~\ref{app:qformat}.

\subsection{Quality Control}
To assess annotation quality, sampled instances from each of the four components — quiz filtering, evidence-scene mapping, triple extraction, and style transfer — were independently re-annotated by a reviewer who did not participate in the initial annotation. Agreement with the initial annotation was 99\% for quiz filtering (n=100), 97\% for evidence-scene mapping (n=100), 94\% for triple extraction (n=50), and 96.5\% for style transfer (n=200).

\section{GroupChat-Mem}
In long-running group chats, evidence relevant to a query may be distributed across sessions, while other queries hinge on the precise wording of a particular exchange. Retrieving only raw dialogue can miss distributed evidence, whereas relying solely on compressed records can omit conversational detail~\citep{han2025rag}. To meet both requirements, we develop GroupChat-Mem, a dual-view memory baseline that maintains an automatically induced relational memory alongside a raw session repository, retrieves evidence from both, and uses them jointly for response generation (Figure~\ref{figure3}).

\definecolor{proposedrow}{RGB}{238,244,250}
\newcommand{\std}[1]{{\scriptsize (#1)}}

\begin{table*}[htbp]
\centering
\renewcommand{\arraystretch}{1.1}
\resizebox{\textwidth}{!}{%
\begin{tabular}{ll ccc ccc ccc}
\toprule
\multicolumn{2}{c}{\multirow{2}{*}{Memory System}}
& \multicolumn{3}{c}{Friends}
& \multicolumn{3}{c}{The Big Bang Theory}
& \multicolumn{3}{c}{The Office} \\
\cmidrule(lr){3-5}
\cmidrule(lr){6-8}
\cmidrule(lr){9-11}
& & Qwen3.5 & Gemma4 & GPT-OSS
  & Qwen3.5 & Gemma4 & GPT-OSS
  & Qwen3.5 & Gemma4 & GPT-OSS \\
\midrule

\multirow{2}{*}{Reference Settings}
& Random Choice
& 20.00 & 20.00 & 20.00
& 20.00 & 20.00 & 20.00
& 20.00 & 20.00 & 20.00 \\
& Direct QA (No History)
& 21.36 \std{0.57} & 28.72 \std{0.48} & 24.62 \std{1.90}
& 22.53 \std{0.52} & 27.25 \std{0.76} & 29.61 \std{1.44}
& 28.22 \std{0.29} & 32.28 \std{0.39} & 30.34 \std{0.95} \\

\midrule

\multirow{3}{*}{\shortstack{Oracle Memory\\(Upper Bound)}}
& Sessions+TKGs
& 96.70 \std{0.99} & 96.79 \std{0.74} & 93.23 \std{0.95}
& 95.45 \std{0.61} & 95.11 \std{0.59} & 92.84 \std{0.75}
& 91.79 \std{0.32} & 92.91 \std{0.27} & 91.00 \std{0.63} \\
& Sessions
& 77.20 \std{0.29} & 79.15 \std{0.89} & 77.45 \std{1.28}
& 85.13 \std{0.26} & 81.95 \std{1.39} & 85.80 \std{0.61}
& 76.99 \std{0.60} & 78.00 \std{0.74} & 77.96 \std{1.03} \\
& TKGs
& 64.09 \std{1.10} & 64.21 \std{0.77} & 63.07 \std{0.79}
& 63.73 \std{0.87} & 65.26 \std{0.56} & 64.64 \std{0.31}
& 58.24 \std{0.20} & 59.78 \std{0.42} & 60.08 \std{0.26} \\

\midrule
\addlinespace[2pt]
\midrule

\multirow{2}{*}{Simple RAG}
& BM25
& 69.88 \std{0.89} & 69.29 \std{1.79} & 56.22 \std{1.21}
& 69.69 \std{2.66} & 68.90 \std{1.41} & 61.49 \std{1.88}
& 65.79 \std{0.24} & 64.48 \std{0.13} & 59.59 \std{0.40} \\
& OpenAI Embedding
& 70.39 \std{0.76} & 70.18 \std{1.21} & 58.71 \std{0.19}
& 74.91 \std{1.55} & 76.02 \std{1.61} & 67.83 \std{0.69}
& 74.27 \std{1.37} & 74.58 \std{1.06} & 67.01 \std{0.50} \\

\midrule

\multirow{4}{*}{Agentic Memory}
& HippoRAG 2
& 73.18 \std{1.09} & 73.01 \std{1.15} & 60.07 \std{0.81}
& \textbf{78.10} \std{0.36} & 77.43 \std{1.78} & 68.07 \std{1.49}
& 75.13 \std{0.79} & 75.32 \std{0.66} & 68.53 \std{0.85} \\

& Mem0
& 71.15 \std{0.78} & 71.45 \std{1.87} & 62.27 \std{1.42}
& 73.87 \std{1.34} & 71.35 \std{0.19} & 64.89 \std{0.71}
& 68.77 \std{1.03} & 69.88 \std{0.37} & 66.04 \std{0.70} \\

& A-MEM
& 65.23 \std{0.96} & 69.20 \std{0.37} & 57.95 \std{0.89}
& 69.57 \std{0.12} & 70.19 \std{0.33} & 61.82 \std{0.31}
& 68.06 \std{0.83} & 69.32 \std{0.38} & 62.69 \std{0.28} \\

\rowcolor{proposedrow}
& \textbf{GroupChat-Mem}
& \textbf{77.83} \std{2.48}
& \textbf{77.58} \std{1.18}
& \textbf{64.81} \std{1.17}
& 77.52 \std{2.48}
& \textbf{78.92} \std{0.85}
& \textbf{70.27} \std{0.56}
& \textbf{78.68} \std{0.98}
& \textbf{77.06} \std{0.34}
& \textbf{69.00} \std{0.34} \\

\bottomrule
\end{tabular}%
}
\vspace{-2mm}
\caption{Main experimental results on GroupChat-Bench (New Names setting), reported as the mean and standard deviation over three runs with different sampled question sets. The best results are highlighted in \textbf{bold}.}
\vspace{-2mm}
\label{tab:main_results}
\end{table*}

\paragraph{Memory Construction from Dialogues}
As the dialogue $\mathcal{D} = \{S_1, S_2, ..., S_N\}$ progresses, GroupChat-Mem maintains a relational memory $\mathcal{R}$ and a raw session repository $\mathcal{S}$.

For $\mathcal{R}$, an LLM performs Open Information Extraction (OpenIE) \citep{banko2007openie}, producing records of the form [subject, relation, object]. The extraction is open-schema and does not rely on predefined entity types or relation labels. Extraction is triggered at two points: when a query arrives during session $S_i$, the LLM extracts records from the partial session $S_{i,j}$ observed up to that turn, and when the session ends, it extracts once more from the complete session $S_i$. Each extraction covers the entire (partial) session and replaces any earlier extraction for $S_i$, so that records are revised as later utterances resolve references or clarify prior statements. The extracted records form a graph-structured memory in which subjects and objects serve as nodes and relations as edges, each timestamped with its source session. Shared nodes connect records from different sessions, linking related evidence across session boundaries.
In parallel, each original session $S_i$ is stored in $\mathcal{S}$, preserving exact wording, local discourse context, and conversational nuances that relational extraction may omit (see Algorithm~\ref{alg:groupchat_mem_memory_construction} in Appendix~\ref{app:groupchat_mem_algorithm} for details).

Note that $\mathcal{R}$ is induced online from the observed dialogue only and is fully independent of the TKG used for benchmark construction — it accesses none of the TKG's relation schema, verified triples, or annotations.

\noindent\textbf{Memory Retrieval and Response Generation}
Given a user query, \memname retrieves evidence from both memory components, $\mathcal{R}$ and $\mathcal{S}$. 
This parallel retrieval combines structured relational information from $\mathcal{R}$ with detailed conversational context from $\mathcal{S}$, without committing to a single retrieval source in advance.

For retrieval from $\mathcal{R}$, \memname first prompts the LLM to identify the top-$n$ query-relevant nodes and constructs a two-hop neighborhood around each selected node. The records contained in these neighborhoods are then linearized and ranked according to their cosine similarity to the query~\citep{karpukhin2020dense}, using OpenAI embeddings~\cite{openai_new_embedding_models_2024}. The top-$k_r$ records are retained as the initial relational evidence.

For retrieval from $\mathcal{S}$, each dialogue session is treated as a retrieval unit. Relevant sessions are retrieved using a combination of lexical and semantic matching. Specifically, BM25~\cite{robertson2009probabilistic} captures keyword-level relevance, while embedding-based cosine similarity captures semantic relevance. \memname retrieves $k_s$ sessions in total, with half selected according to their BM25 scores and the other half according to their embedding-based cosine similarity to the query. The retrieved sessions serve as raw session evidence for response generation.

Using the evidence retrieved from both memory views, \memname prompts the LLM to generate an initial response and assess whether the query can be answered with sufficient support. 
If the retrieved evidence is sufficient, the response is returned directly. Otherwise, \memname initiates a fine-grained traversal over the induced relational memory.

Among the nodes appearing in the initially retrieved records, \memname prompts the LLM to select the node most relevant to the query. Starting from the selected node, \memname expands the relational neighborhood one hop at a time. After each expansion, the LLM attempts to answer the query using the accumulated relational evidence and assesses whether the resulting answer is sufficiently supported. The traversal terminates when a supported answer can be generated or when the maximum hop limit is reached. If the available dialogue evidence remains insufficient, \memname returns ``I don't know.''
The complete retrieval procedure is summarized in Algorithm~\ref{alg:groupchat_memory_retrieval_qa} (Appendix~\ref{app:groupchat_mem_algorithm}).

\section{Experiments}
\subsection{Experimental Setting}
\label{expsetting}

In our experiments, we used the multiple-choice format to efficiently and accurately assess the agent's dialogue understanding capabilities. 
To fairly compare GroupChat-Mem with other baselines, we used Qwen3.5-35B-A3B~\cite{qwen2026qwen35_35b_a3b}, Gemma4-26B-A4B~\cite{google2026gemma4_26b_a4b}, and GPT-OSS 20B~\cite{openai2025gptoss120bgptoss20bmodel} as the LLM backbones. Detailed experimental settings, including the hyperparameters used in our experiments, are provided in Appendix~\ref{app:avgtime}.

\subsection{Baselines}
\label{sec5.2}
We compare GroupChat-Mem against three representative agentic memory systems, along with two simple RAG baselines. For all baselines, we used the official implementations and tuned their hyperparameters and prompts for evaluation on GroupChat-Bench.

\noindent\textbf{BM25 and OpenAI Embedding} serve as representative simple RAG baselines, using sparse lexical retrieval and dense semantic retrieval, respectively.

\noindent\textbf{HippoRAG 2}~\cite{gutierrez2025rag} is a graph-based retrieval framework that constructs associative memory over textual evidence and retrieves query-relevant passages through graph search.

\noindent\textbf{Mem0}~\cite{chhikara2025mem0} maintains long-term agent memory by extracting, updating, and deleting compact memory items from ongoing interactions, rather than storing the full dialogue history as-is. 

\noindent\textbf{A-MEM}~\cite{xu2026mem} constructs an agentic memory system by converting past interactions into memory notes and dynamically linking them according to semantic and contextual relevance.

\subsection{Results}
\label{sec5.3}
As shown in Table~\ref{tab:main_results}, \memname achieves the best performance in 8 of 9 settings, outperforming the strongest baseline by up to 4.7 points. Nevertheless, all memory systems remain below 80, leaving a gap of roughly 20 points to the Oracle upper bound (Sessions+TKGs). This gap highlights that retrieving relevant evidence from long group-chat histories remains the key bottleneck.

\paragraph{Oracle Memory}
LLMs achieve scores of 91--96 when given both the evidence sessions for each question and the corresponding TKGs. Interestingly, performance drops substantially when either representation is removed, even though the evidence sessions alone contain sufficient information to answer the questions. This suggests that raw multi-party dialogue remains difficult for LLMs to interpret and that explicit structural representations can substantially aid reasoning.

\begin{table}[t]
\centering
\resizebox{\columnwidth}{!}{%
\footnotesize
\begin{tabular}{lccc}
\toprule
\textbf{Memory System} & \textbf{New Names} & \textbf{Original Names} & \textbf{Swapped Names} \\
\midrule
BM25
& 65.13 & 66.29 (\textcolor{blue}{+1.16}) & 61.97 (\textcolor{red}{-3.16}) \\
OpenAI Embedding
& 66.43 & 68.01 (\textcolor{blue}{+1.58}) & 65.20 (\textcolor{red}{-1.23}) \\
HippoRAG 2
& 68.75 & 69.97 (\textcolor{blue}{+1.22}) & 66.77 (\textcolor{red}{-1.98}) \\
Mem0
& 68.29 & 72.60 (\textcolor{blue}{+4.31}) & 66.97 (\textcolor{red}{-1.32}) \\
A-MEM
& 64.13 & 66.62 (\textcolor{blue}{+2.49}) & 62.00 (\textcolor{red}{-2.13}) \\
GroupChat-Mem
& 73.41 & 75.18 (\textcolor{blue}{+1.78}) & 71.57 (\textcolor{red}{-1.84}) \\
\bottomrule
\end{tabular}%
}
\vspace{-2mm}
\caption{Results on GroupChat-Bench (Friends) under different
character-name settings (New: Joey $\rightarrow$ John; Swapped:
Joey $\leftrightarrow$ Monica), averaged across all backbone models.
Deltas are relative to New Names.}
\vspace{-2mm}
\label{tab:data_type_result}
\end{table}

\paragraph{Impact of Pretrained Knowledge}
Because LLMs may have prior knowledge of the TV shows used in our
benchmark, we replace character names with new names by default to
reduce reliance on memorized information. Nevertheless, in Direct QA
without dialogue history, models scored above the random-choice
baseline (20\%) and reached up to 32.3\%, indicating that name replacement
alone is insufficient. To further examine the effect of character
names, we compare two additional settings: Original Names and the adversarial
Swapped Names (\textit{e.g.}, Joey $\leftrightarrow$ Monica). As shown in
Table~\ref{tab:data_type_result}, performance improves with Original
Names but declines below the New Names default with Swapped Names.
This suggests that swapped names not only reduce the usefulness of
memorized knowledge but also make it actively misleading by
conflicting with the dialogue context.

\begin{figure}[t]
    \centering
    \begin{center}
    \includegraphics[width=\linewidth]{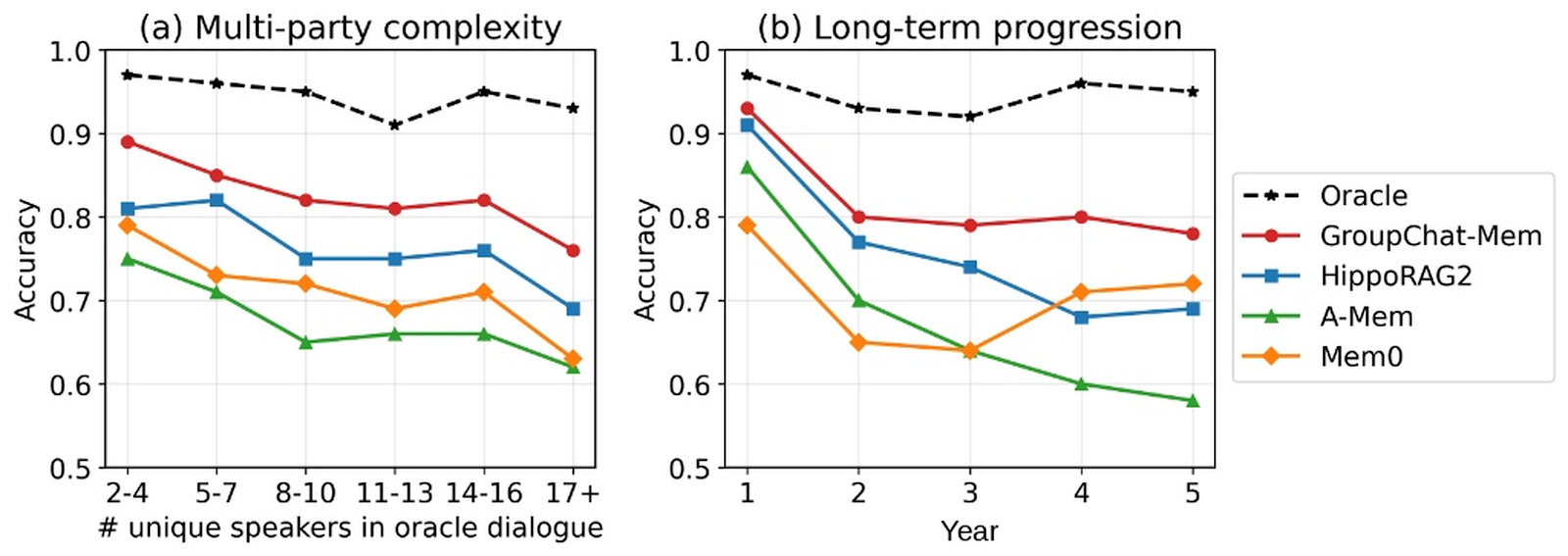}
    \end{center}
    \vspace{-4mm}
    \caption{Analysis on the performance of each method with respect to (a) multi-party complexity and (b) long-term progression of the given dialogue.}
    \vspace{-3.5mm}
    \label{figure4}
\end{figure}

\paragraph{Challenges in Long-Term Multi-Party Dialogue Understanding}
We analyzed baseline performance by the number of participants in the evidence sessions and by dialogue length. As shown in Figure~\ref{figure4}(a), performance decreases as the number of participants increases, demonstrating the difficulty of tracking information and relationships across multiple speakers. Figure~\ref{figure4}(b) shows a similar decline as dialogue history grows, indicating that retrieving and integrating relevant evidence becomes increasingly challenging over long-term interactions. However, GroupChat-Mem degrades more gracefully than the other methods, highlighting its robustness to extended conversational histories. These results identify speaker multiplicity and dialogue length as key challenges in long-term group-chat understanding and demonstrate the advantage of GroupChat-Mem in the latter setting.

\paragraph{Further Analyses} We additionally examine the effect of question format, the contribution of each memory component, and the behavior of the induced relational memory. 
First, evaluation format matters: GroupChat-Mem accuracy drops by 13–29 points across backbones when moving from labeled MCQ (choices given as explicit options) to open-ended evaluation (\textit{e.g.}, 77.8→64.5 for Qwen3.5; Appendix~\ref{app:q_format_performance}). Because answer choices bound the output space, the two formats probe complementary abilities: labeled MCQ offers reliable automatic scoring of discriminative understanding, motivating its use as our default, while open-ended evaluation measures generative recall and leaves considerable headroom.
Second, component ablations (Appendix~\ref{app:ablation_component}) show that both memory views are necessary, though to different degrees: removing the session repository is most damaging (-19.8 points on average), as compressed records omit the precise wording that many queries hinge on, while removing the relational memory costs 4.6 points, as evidence linked across distant sessions can no longer be resolved. Disabling the iterative graph traversal further degrades performance by 1.5 points, confirming its role in incrementally gathering additional evidence when initial retrieval is insufficient. 
The appendix also reports a breakdown of accuracy by question type
(Appendix~\ref{app:q_type_perf}), the distribution of gold answer
choices (Appendix~\ref{app:true_answer_option_ratio}), and qualitative
examples showing how GroupChat-Mem retrieves evidence across session
boundaries (Appendix~\ref{app:qualitative_analysis}).

\section{Conclusion}
We introduced GroupChat-Bench, a benchmark for evaluating long-term multi-party dialogue understanding over chronologically accumulating group-chat histories. The benchmark tests whether LLM agents can track participant-specific information, evolving relationships, and contextual changes across extended interactions. We also proposed GroupChat-Mem, which combines relational memory with a raw session repository. Experiments reveal limitations in existing memory systems, while GroupChat-Mem provides a stronger and more robust baseline for long dialogue histories.

\section*{Limitations}
\paragraph{Scripted dialogue as a proxy for group chats.}
Ideally, GroupChat-Bench would be built from naturally occurring group
communication, such as corporate Slack channels or workplace meeting
transcripts. However, such data is difficult to access and release:
it raises serious privacy and consent concerns for the participants,
and is typically treated as proprietary corporate assets. We therefore
use TV scripts as a proxy, since they provide human-authored,
long-horizon multi-party dialogue with persistent character
relationships, free of these constraints. To partially compensate for
the lack of real-world data, we selected shows that cover diverse
everyday contexts, including interpersonal relationships
(\textit{Friends}), academic settings (\textit{The Big Bang Theory}),
and workplace interactions (\textit{The Office}), though they remain
constrained to narrative-driven dialogue. These scripts retain key
conversational phenomena such as fragmented turn-taking, pile-on reactions, and topic shifts across speakers (Figure~\ref{figure1}),
but they do not capture some surface characteristics of text-based
messaging, such as typos, emojis, or asynchronous short reactions.
We expect the core challenges evaluated by GroupChat-Bench
(multi-party grounding, temporal persistence, and evidence distributed
across noisy exchanges) to transfer to such settings, though extending
the benchmark with naturally occurring (appropriately consented) chat
data is a natural next step.

\paragraph{Data contamination.}
Since the source shows likely appear in pretraining corpora, we
mitigate contamination by replacing character names with unseen ones
(the New Names setting), adversarially swapping names among characters
(the Swapped Names setting), and including unanswerable questions.
We further quantify residual leakage directly via the Direct QA
baseline (Section~\ref{sec5.3}). While these measures make memorized knowledge
unreliable or even misleading for solving the benchmark, they cannot rule out that models exploit memorized plot structure.
Constructing a comparable benchmark from dialogue guaranteed to be
unseen during pretraining remains an open challenge for the field.

\section*{Ethics Statement}
\paragraph{Data sources and copyright.}
GroupChat-Bench is built from fan-transcribed scripts of three TV 
shows, obtained from publicly available Kaggle datasets, and from 
fan-created quizzes on FunTrivia. The underlying shows are 
copyrighted works; our use and release are strictly limited to 
non-commercial academic research, following established practice in 
prior dialogue benchmarks constructed from the same transcripts 
\citep{yang-choi-2019-friendsqa, poria-etal-2019-meld}. The released dialogue differs from the 
original scripts: all non-verbal descriptions and production 
metadata are removed. Fan-quiz questions are likewise filtered and rewritten 
rather than reproduced verbatim (\S3.1.2, \S3.1.5). We release the 
benchmark under a research-only license and will promptly remove 
any content upon request from the relevant rights holders.

\section*{Use of AI Assistants}
We used an LLM-based writing assistant solely to polish the writing of
this paper, i.e., for grammar correction and paraphrasing of
author-written text. It was not used to generate scientific content,
ideas, analyses, or citations, and all suggestions were reviewed and
edited by the authors, who take full responsibility for the final
content. Separately, GPT-4 was used as a component of the dataset
construction pipeline as a research object.

\section*{Acknowledgments}
This work was supported by the Institute of Information \& Communications Technology Planning \& Evaluation (IITP) grants (RS-2019-II190075, RS-2024-00338140), the InnoCORE Program (N10250156), and the National Research Foundation of Korea (NRF) grant (NRF-2020H1D3A2A03100945, RS-2024-00333484), funded by the Korean government (MSIT).

\bibliography{custom}

\appendix

\section{Date Assignment}
\label{app:daterule}
We first extracted elements from the scripts that could indicate dates (\textit{e.g.}, Valentine's Day, Christmas Eve). Then, we reviewed the scripts again to analyze the relative timing of the sessions. For example, if there is a line mentioning that Chandler broke up with his girlfriend two days ago, we annotated the session where he broke up with his girlfriend as occurring two days prior to the mentioned session. Next, while watching each episode, we pinpointed sessions where the dates might have changed by observing whether the characters' outfits changed between sessions. Finally, we assigned a specific date to each session based on the actual broadcast date of the episode, adjusting for the relative differences in dates and events such as Christmas.

\section{Question Generation Based on Fan Quizzes}
\label{app:fangen}
For each scene $s_{i, k}$ from episode $p_i$ in $Script_{pre}$, we define the set of answerable questions as $FanA_{i, k}$ and the set of unanswerable questions as $FanU_{i, k}$. The process of generating questions based on fan quizzes is as follows.

First, we collected quizzes for each season and episode of Friends, The Big Bang Theory, and The Office from the FunTrivia website. For each episode $p_i$ in $Script_{pre}$, we used GPT-4 to determine if the crawled questions ${CrQ}_i = \{q_{i,0}, q_{i, 1}, ..., q_{i, l}\}$ could be answered using only $p_i$. If a question $q_{i, m}$ could be answered, \mbox{GPT-4} identified the scenes $ES_{i, m}$ that provide evidence for the answer, compiling them into $Q_i = \{(q_{i, m}, ES_{i, m})\}_{m=0}^l$. Subsequently, the authors reviewed each $ES_{i, m}$, made necessary corrections, and annotated whether a single scene from $ES_{i, m}$ was sufficient to answer $q_{i, m}$ or if multiple scenes were needed to be considered simultaneously. For each $s_{i, k}$ within $p_i$, we assessed the answerability of the questions in $Q_i$. 

For each $s_{i,k}$, if a question $q_{i, m}$ could be answered using just one scene, and $s_{i, k}$ occurs after the initial appearance of the main character in $ES_{i, m}$, we included $q_{i, m}$ in $FanA_{i, k}$. This ensures that the main character had adequate exposure to the relevant evidence. Additionally, for questions requiring verification across multiple scenes, if the main character appears in all $ES_{i, m}$ scenes and $s_{i, k}$ occurs after the last scene of $ES_{i, m}$, we included $q_{i, m}$ in $FanA_{i, k}$. If the main character does not appear in any of the $ES_{i, m}$ scenes, $q_{i, m}$ was included in $FanU_{i, k}$ since the main character has not experienced any evidence to answer the question. The rest are not included in the dataset as it is unclear whether they are answerable per scene.
Additionally, to generate questions that require long-term memory, we added the most recent date of the evidence scenes for each question.

\section{Question Generation Based on a Temporal Knowledge Graph}
\label{app:tkggen}

\subsection{Relations}
\label{app:tkgrel}
We used the following 32 relations: `age', `alumni', `boss', `boyfriend', `brother', `client', `date of birth', `dating with', `ex-boyfriend', `ex-fiance', `ex-fiancee', `ex-girlfriend', `ex-husband', `ex-roommate', `ex-wife', `father', `fiance', `fiancee', `girlfriend', `hometown', `husband', `job', `major', `mother', `neighbor', `pet', `place of birth', `place of work', `roommate', `sister', `subordinate', `wife'.

\subsection{Question Templates and Generated Questions}
\label{app:tkgqexam}
Templates for one-hop questions are provided in Table~\ref{app:question_onehop_without_time_templates} and Table~\ref{app:question_onehop_time_templates}. The former contains templates without temporal information, while the latter includes templates with temporal details. Since relations like ``brother'' and ``sister'' remain constant over time, questions about these relations do not require temporal information. Hence, no temporal templates were created for them. In Table~\ref{app:question_onehop_time_templates}, ``on \{time\}'' is used, but \{time\} can be not only the full date (year, month, and day) but also just the year and month, or even just the year. In these cases, ``in \{time\}'' is used.

The templates for two-hop questions are available in Table~\ref{app:question_twohop_templates}.
These templates incorporate temporal information. 
To frame questions in the present tense, adjust the verbs to the present tense and remove the temporal information, following the approaches demonstrated in Table~\ref{app:question_onehop_without_time_templates} and Table~\ref{app:question_onehop_time_templates}.

\section{Character Style Transfer}
\label{app:styletrans}
Table~\ref{app:question_styletransfer} shows the results of the character style transfer for three selected questions. 
To make the questions sound more natural and conversational, we prepended each one with ``By the way,''. This helps them blend seamlessly into the flow of the conversation.
The table shows how each question appears when rephrased in the style of various characters. The `Default' setting is applied when the question is asked by a character who is not a recurring character of the TV show.



\section{Experimental Setting}
\label{app:avgtime}

\subsection{Algorithm}
\label{app:groupchat_mem_algorithm}

This section explains the full GroupChat-Mem algorithm in detail.
Algorithm~\ref{alg:groupchat_mem_memory_construction} describes how the two
memories, $\mathcal{R}$ and $\mathcal{S}$, are constructed. Extraction always
takes the \emph{complete} partial session $S_{i,j}$ as input rather than the
latest utterance alone, so references resolved later in the session are
reflected in the resulting records. A query posed at turn $j$ is answered over
the point-in-time views $\mathcal{R}^{(i,j)}$ and $\mathcal{S}^{(i,j)}$, which
add the current partial session to what was committed before $S_i$; records
from different sessions are connected, linking evidence
across session boundaries.

Algorithm~\ref{alg:groupchat_memory_retrieval_qa} details retrieval and
response generation. When the initial answer over the parallel evidence of
Stage~1 is ``I don't know.'', Stage~2 repeatedly asks the LLM which entity to
look up next and expands that entity's relational neighborhood, re-answering
after each expansion. The traversal stops when an answer is produced, when the
LLM reports no further entity to examine, or when the hop limit $H$ is reached,
in which case GroupChat-Mem returns ``I don't know.''

\begin{algorithm}[htbp]
\small
\caption{GroupChat-Mem: Memory Construction}
\label{alg:groupchat_mem_memory_construction}
\KwIn{Dialogue stream $\mathcal{D}=\{S_1,\dots,S_N\}$;\\
      timestamps $\{t_i\}_{i=1}^{N}$; LLM $\mathcal{M}$;\\
      query turn $j$ within each session $S_i$}
\KwOut{Committed memories $\mathcal{R},\mathcal{S}$; query-time views
       $\mathcal{R}^{(i,j)},\mathcal{S}^{(i,j)}$}
$\mathcal{R}\gets\emptyset$,\ \ $\mathcal{S}\gets\emptyset$\;
\For{$i=1$ \KwTo $N$}{
  \tcc{point-in-time view at query turn $j$}
  $E_{i,j}\gets\textsc{OpenIE}(\mathcal{M},S_{i,j})$ \tcp*{complete partial session}
  $\mathcal{R}^{(i,j)}\gets\mathcal{R}\cup\{(s,r,o,t_i)\mid(s,r,o)\in E_{i,j}\}$\;
  $\mathcal{S}^{(i,j)}\gets\mathcal{S}\cup\{(S_{i,j},t_i)\}$\;
  \tcc{commit once the session ends}
  $E_i\gets\textsc{OpenIE}(\mathcal{M},S_i)$\;
  $\mathcal{R}\gets\mathcal{R}\cup\{(s,r,o,t_i)\mid(s,r,o)\in E_i\}$\;
  $\mathcal{S}\gets\mathcal{S}\cup\{(S_i,t_i)\}$ \tcp*{verbatim}
}
\Return $\mathcal{R},\mathcal{S}$\;
\end{algorithm}

\begin{algorithm}[htbp]
\small
\caption{GroupChat-Mem: Retrieval and Response Generation}
\label{alg:groupchat_memory_retrieval_qa}
\KwIn{Query $q$ at turn $j$ of $S_i$; views $\mathcal{R}^{(i,j)},\mathcal{S}^{(i,j)}$ (Alg.~\ref{alg:groupchat_mem_memory_construction});\\
      LLM $\mathcal{M}$; embedder $f(\cdot)$; params $n,k_r,k_s$; hop limit $H$}
\KwOut{Answer $a$}
\tcc{Stage 1: parallel retrieval}
$V\gets\textsc{SelectNodes}(\mathcal{M},q,\mathcal{R},n)$\;
$C\gets\bigcup_{v\in V}\textsc{TwoHopNbhd}(\mathcal{R},v)$
  \tcp*{if $V=\varnothing$: records matching $q$'s timestamp}
$E_{\mathcal{R}}\gets\textsc{Top}_{k_r}\!\big(e\in C;\cos(f(\textsc{Lin}(e)),f(q))\big)$\;
$\begin{aligned}[t]
  E_{\mathcal{S}}\gets{}&\textsc{Top}_{k_s/2}\big(\mathcal{S};\mathrm{BM25}(q,\cdot)\big)\\
  &\cup\ \textsc{Top}_{k_s/2}\big(\mathcal{S};\cos(f(\cdot),f(q))\big)
\end{aligned}$\;
$a\gets\mathcal{M}(q,E_{\mathcal{R}},E_{\mathcal{S}})$\;
\lIf{$a\neq\textit{``I don't know.''}$}{\Return $a$}
\tcc{Stage 2: entity-guided traversal}
$F\gets\varnothing$,\ \ $V_{\text{vis}}\gets\varnothing$,\ \ \textit{exhausted}$\gets$\textbf{true}\label{lin:traversal_start}\;
\For{$h=1$ \KwTo $H$}{
  $o\gets\textsc{NextLookup}(\mathcal{M},q,E_{\mathcal{R}},E_{\mathcal{S}},F,V_{\text{vis}})$
    \tcp*{entity / \textsc{Unknown} / \textsc{Idk}}
  \lIf{$o\in\{\textsc{Idk},\textsc{Unknown}\}$}{\Return \textit{``I don't know.''}}
  $v\gets\textsc{Resolve}(o,\textsc{Nodes}(\mathcal{R}))$\;
  \lIf{$v=\bot$ \textbf{or} $v\in V_{\text{vis}}$}{\textit{exhausted}$\gets$\textbf{false}; \textbf{break}}
  $\Delta\gets\textsc{OneHop}(\mathcal{R},v)\setminus(E_{\mathcal{R}}\cup F)$\;
  \lIf{$\Delta=\varnothing$}{\textit{exhausted}$\gets$\textbf{false}; \textbf{break}}
  $V_{\text{vis}}\gets V_{\text{vis}}\cup\{v\}$,\ \ $F\gets F\cup\Delta$\;
  $a\gets\mathcal{M}(q,E_{\mathcal{R}}\cup F,E_{\mathcal{S}})$\;
  \lIf{$a\neq\textit{``I don't know.''}$}{\Return $a$}
}
\lIf{\textit{exhausted}}{\Return \textit{``I don't know.''}}
\Return $\mathcal{M}(q,E_{\mathcal{R}},E_{\mathcal{S}})$\label{lin:traversal_end}\;
\end{algorithm}

\subsection{Hyperparameters}
\label{app:hyperparam}
The value and meaning of each hyperparameter used for the experiments are provided in Table~\ref{tab:hyperparam}, Table~\ref{tab:hyperparam_amem}, Table~\ref{tab:hyperparam_mem0}, Table~\ref{tab:hyperparam_hipporag2}, and Table~\ref{tab:hyperparam_general}.

\begin{table}[h!]
\centering
\renewcommand{\arraystretch}{1.1}
\resizebox{\columnwidth}{!}{%
\begin{tabular}{lcl}
\toprule
Hyperparameter & Value & Meaning \\
\midrule
$k_{\text{r}}$ & 250 & \# retrieved relational evidence \\
$k_{\text{s}}$ & 40 & \# retrieved sessions \\
$n$ & 3 & \# query-relevant nodes extracted by LLMs \\
\bottomrule
\end{tabular}%
}
\vspace{-2mm}
\caption{Hyperparameters used for Groupchat-Mem experiments. Detailed explanation on hyperparameter ablation is available on Appendix~\ref{app:ablation_hyperparam}.}
\vspace{-2mm}
\label{tab:hyperparam}
\end{table}

\begin{table}[h!]
\centering
\renewcommand{\arraystretch}{1.1}
\resizebox{\columnwidth}{!}{%
\begin{tabular}{lcl}
\toprule
Hyperparameter & Value & Meaning \\
\midrule
$k_{\text{neighbors}}$      & 5      & \# nearest notes retrieved for memory update \\
turns\_per\_note            & 5      & \# utterances used to construct each note \\
num\_retrieved            & 100      & \# notes retrieved for each QA \\
max\_hops                 & 6      & \# max iterations during the entity-guided traversal\\
\bottomrule
\end{tabular}%
}
\caption{Hyperparameters used for A-MEM experiments.}
\label{tab:hyperparam_amem}
\end{table}

\begin{table}[h!]
\centering
\renewcommand{\arraystretch}{1.1}
\resizebox{\columnwidth}{!}{%
\begin{tabular}{lcl}
\toprule
Hyperparameter & Value & Meaning \\
\midrule
turns\_per\_batch           & 5      & \# utterances per fact-extraction \\
top\_s                      & 10     & \# memories used for each update \\
\bottomrule
\end{tabular}%
}
\caption{Hyperparameters used for Mem0 experiments.}
\label{tab:hyperparam_mem0}
\end{table}

\begin{table}[h!]
\centering
\renewcommand{\arraystretch}{1.1}
\resizebox{\columnwidth}{!}{%
\begin{tabular}{lcl}
\toprule
Hyperparameter & Value & Meaning \\
\midrule
num\_retrieved              & 40     & \# passages retrieved for QA \\
synonymy\_sim\_threshold    & 0.8    & Cosine-similarity threshold \\
synonymy\_topk              & 2047   & Max \# synonym neighbors \\
linking\_top\_k             & 5      & \# phrase nodes seeded for PPR \\
passage\_node\_weight       & 0.05   & Relative weight of passage nodes \\
damping                     & 0.5    & PPR damping (restart) factor \\
use\_filter                 & True   & Applying the LLM fact-filter \\
\bottomrule
\end{tabular}%
}
\caption{Hyperparameters used for HippoRAG 2 experiments.}
\label{tab:hyperparam_hipporag2}
\end{table}

\begin{table}[h!]
\centering
\renewcommand{\arraystretch}{1.1}
\resizebox{\columnwidth}{!}{%
\begin{tabular}{lcl}
\toprule
Hyperparameter & Value & Meaning \\
\midrule
\multicolumn{3}{l}{\textit{Decoding (client sampling)}} \\
temperature (Qwen, Gemma) & 0.6 & Sampling temperature \\
temperature (GPT-OSS) & 1.0 & Sampling temperature \\
top\_p (Qwen, Gemma) & 0.95 & Nucleus sampling probability mass \\
top\_p (GPT-OSS)  & 1.0 & Nucleus sampling probability mass \\
max\_tokens        & 16384  & Max generated tokens per LLM call \\
thinking\_token\_budget & 10000 & Reasoning-token budget for QA \\
reasoning\_effort  & high & Reasoning level for QA \\
\midrule
\multicolumn{3}{l}{\textit{vLLM serving}} \\
max-model-len            & 65536 & Max context length served \\
gpu-memory-utilization   & 0.90 & Fraction of reserved GPU memory \\
max-num-seqs             & 256 & Max concurrent decoding sequences \\
\bottomrule
\end{tabular}%
}
\caption{General serving and decoding hyperparameters shared across all experiments. Qwen/Gemma use \texttt{thinking\_token\_budget}; GPT-OSS is harmony-format and instead uses \texttt{reasoning\_effort}.}
\label{tab:hyperparam_general}
\end{table}

\subsection{Question Format}
\label{app:qformat}
\dataname is a dataset that includes pairs of questions, answers, and choices. The questions are available in three formats: template-based multiple-choice, natural language multiple-choice, and open-ended. Users can choose any of these formats to evaluate the agent's performance.

First, we provide multiple-choice questions in both template and natural language formats. For example, a template-based question might be, ``Who was going out with Paul in September 1994?'' with choices ``(A) Emily, (B) Monica, (C) Ryan, (D) Rachel, (E) I don’t know''. In contrast, the same question in natural language format could be phrased as, ``Who was going out with Paul in September 1994? Was it Emily, Monica, Ryan, Rachel, or do you not know?''

Additionally, we offer the option to ask questions in an open-ended format (\textit{e.g.}, ``Who was going out with Paul in September 1994?'') without providing answer choices. This approach allows us to evaluate the agent's ability to generate open-ended responses. The open-ended format is particularly useful for fan quiz-based questions, where some answers may require longer responses (\textit{e.g.}, Question: ``Why did Monica and Chandler say they were late getting to the hospital?'' Correct answer: ``Monica went back for her jacket'').

For natural language multiple-choice and open-ended questions, we used a generative reading-comprehension protocol rather than exact multiple-choice label matching. Since the model produces free-form answers, we adopt the official SQuAD v1.1 evaluation~\citep{rajpurkar2016squad}: each answer is normalized (lowercasing, punctuation and article removal, whitespace normalization) and scored against every candidate option using token-level F1 and exact match.

\subsubsection{Choices in Multiple-Choice Questions}
The number of questions based on fan quizzes was significantly smaller than the questions based on the TKG. Thus, 50\% of the questions were intentionally extracted from the fan quiz-based during the simulation. Since each question has five choices, unanswerable questions were set to comprise 20\% of the total to fairly stratify the correct answers.

\subsubsection{Performance on different question formats}
\label{app:q_format_performance}

Table~\ref{tab:q_format_friends_result}-\ref{tab:q_format_theoffice_result} report GroupChat-Mem performance on GroupChat-Bench (New Names) across different question formats.
In most settings, Labeled MCQ achieves the highest performance, while Open-Ended evaluation yields the lowest performance. This suggests that providing answer choices substantially facilitates the task by narrowing the answer space, with explicit option labels offering an additional advantage over choices embedded in natural-language questions.


\begin{table}[h!]
\centering
\renewcommand{\arraystretch}{1.1}
\resizebox{\columnwidth}{!}{%
\begin{tabular}{l ccc}
\toprule
\multirow{2}{*}{Model} & \multicolumn{3}{c}{Friends} \\
\cmidrule(lr){2-4}
& Labeled MCQ & Naturalized MCQ & Open-Ended \\
\midrule
Qwen3.5
& \textbf{77.83} \std{2.48} & 74.62 \std{1.16} & 64.47 \std{0.79} \\
Gemma4
& \textbf{77.58} \std{1.18} & 73.65 \std{1.23} & 58.97 \std{0.89} \\
GPT-OSS
& \textbf{64.81} \std{1.17} & 63.41 \std{0.39} & 45.01 \std{0.51} \\
\bottomrule
\end{tabular}%
}
\vspace{-2mm}
\caption{Performance of GroupChat-Mem on GroupChat-Bench (Friends, New Names) across question formats. Labeled MCQ presents choices explicitly with option labels, whereas Naturalized MCQ incorporates the choices into a natural-language question.}
\vspace{-2mm}
\label{tab:q_format_friends_result}
\end{table}

\begin{table}[h!]
\centering
\renewcommand{\arraystretch}{1.1}
\resizebox{\columnwidth}{!}{%
\begin{tabular}{l ccc}
\toprule
\multirow{2}{*}{Model} & \multicolumn{3}{c}{The BigBang Theory} \\
\cmidrule(lr){2-4}
& Labeled MCQ & Naturalized MCQ & Open-Ended \\
\midrule
Qwen3.5
& \textbf{77.52} \std{2.48} & 76.27 \std{1.29} & 58.26 \std{0.81} \\
Gemma4
& 78.92 \std{0.85} & \textbf{80.04} \std{0.68} & 55.20 \std{1.57} \\
GPT-OSS
& \textbf{70.27} \std{0.56} & 68.82 \std{0.99} & 41.74 \std{0.45} \\
\bottomrule
\end{tabular}%
}
\vspace{-2mm}
\caption{Performance of GroupChat-Mem on GroupChat-Bench (The Big Bang Theory, New Names) across question formats.}
\vspace{-2mm}
\label{tab:q_format_bigbang_result}
\end{table}

\begin{table}[h!]
\centering
\renewcommand{\arraystretch}{1.1}
\resizebox{\columnwidth}{!}{%
\begin{tabular}{l ccc}
\toprule
\multirow{2}{*}{Model} & \multicolumn{3}{c}{The Office} \\
\cmidrule(lr){2-4}
& Labeled MCQ & Naturalized MCQ & Open-Ended \\
\midrule
Qwen3.5
& \textbf{78.68} \std{0.98} & 72.97 \std{0.51} & 52.92 \std{1.05} \\
Gemma4
& \textbf{77.06} \std{0.34} & 76.41 \std{0.66} & 49.51 \std{0.64} \\
GPT-OSS
& \textbf{69.00} \std{0.34} & 68.84 \std{0.85} & 43.20 \std{0.41} \\
\bottomrule
\end{tabular}%
}
\vspace{-2mm}
\caption{Performance of GroupChat-Mem on GroupChat-Bench (The Office, New Names) across question formats.}
\vspace{-2mm}
\label{tab:q_format_theoffice_result}
\end{table}

\subsection{Performance on different question types}
\label{app:q_type_perf}
Table~\ref{tab:q_type_friends} shows the performance of each memory system on GroupChat-Bench (Friends, New Names) per question type. All memory systems show higher performance in Fanquiz questions compared to TKG questions, and GroupChat-Mem outperforms all baselines. 

\begin{table}[h!]
\centering
\renewcommand{\arraystretch}{1.1}
\resizebox{\columnwidth}{!}{%
\begin{tabular}{l ccc}
\toprule
\multirow{2}{*}{Memory System} & \multicolumn{3}{c}{Friends} \\
\cmidrule(lr){2-4}
& Fanquiz & TKG & Overall \\
\midrule
HippoRAG2
& 75.21 \std{0.45} & 62.30 \std{1.90} & 68.75 \std{0.97} \\
A-MEM
& 72.73 \std{0.79} & 55.53 \std{0.66} & 64.13 \std{0.36} \\
Mem0
& 73.58 \std{1.17} & 63.00 \std{1.5} & 68.29 \std{1.23} \\
GroupChat-Mem
& \textbf{79.33} \std{1.21} & \textbf{67.48} \std{1.90} & \textbf{73.41} \std{1.51} \\
\bottomrule
\end{tabular}%
}
\vspace{-2mm}
\caption{Performance of each memory system on GroupChat-Bench (Friends, New Names) per question type.}
\vspace{-2mm}
\label{tab:q_type_friends}
\end{table}

\subsection{The distribution of correct options}
\label{app:true_answer_option_ratio}

Table~\ref{tab:true_answer_option_friends}-\ref{tab:true_answer_option_theoffice} show the count of correct answers per option for each QA set. The correct answers are evenly distributed across the 5 options: (A), (B), (C), (D), and (E).

\begin{table}[h!]
\centering
\renewcommand{\arraystretch}{1.1}
\resizebox{0.7\columnwidth}{!}{%
\begin{tabular}{c ccccc}
\toprule
\multirow{2}{*}{QA set} & \multicolumn{5}{c}{Friends} \\
\cmidrule(lr){2-6}
& (A) & (B) & (C) & (D) & (E) \\
\midrule
QA set 1
& 157 & 157 & 158 & 158 & 158 \\
QA set 2 &
157 & 158 & 157 & 158 & 158 \\
QA set 3
& 158 & 157 & 157 & 158 & 158 \\
All & 472 & 472 & 472 & 474 & 474 \\
\bottomrule
\end{tabular}%
}
\vspace{-2mm}
\caption{The number of correct answers per option in \textbf{Friends} QA sets.}
\vspace{-2mm}
\label{tab:true_answer_option_friends}
\end{table}

\begin{table}[h!]
\centering
\renewcommand{\arraystretch}{1.1}
\resizebox{0.7\columnwidth}{!}{%
\begin{tabular}{c ccccc}
\toprule
\multirow{2}{*}{QA set} & \multicolumn{5}{c}{The Big Bang Theory} \\
\cmidrule(lr){2-6}
& (A) & (B) & (C) & (D) & (E) \\
\midrule
QA set 1
& 161 & 161 & 161 & 162 & 160 \\
QA set 2 &
161 & 162 & 161 & 161 & 160 \\
QA set 3
& 161 & 162 & 161 & 161 & 160 \\
All & 483 & 485 & 483 & 484 & 480 \\
\bottomrule
\end{tabular}%
}
\vspace{-2mm}
\caption{The number of correct answers per option in \textbf{The Big Bang Theory} QA sets.}
\vspace{-2mm}
\label{tab:true_answer_option_bigbang}
\end{table}

\begin{table}[h!]
\centering
\renewcommand{\arraystretch}{1.1}
\resizebox{0.7\columnwidth}{!}{%
\begin{tabular}{c ccccc}
\toprule
\multirow{2}{*}{QA set} & \multicolumn{5}{c}{The Office} \\
\cmidrule(lr){2-6}
& (A) & (B) & (C) & (D) & (E) \\
\midrule
QA set 1
& 469 & 469 & 470 & 470 & 469 \\
QA set 2 &
469 & 470 & 469 & 470 & 469 \\
QA set 3
& 469 & 469 & 470 & 470 & 469 \\
All & 1407 & 1408 & 1409 & 1410 & 1407 \\
\bottomrule
\end{tabular}%
}
\vspace{-2mm}
\caption{The number of correct answers per option in \textbf{The Office} QA sets.}
\vspace{-2mm}
\label{tab:true_answer_option_theoffice}
\end{table}

\section{Ablation Study}
\label{app:ablation}

\subsection{Hyperparameter Ablation}
\label{app:ablation_hyperparam}

To determine the values of each hyperparameter used in Groupchat-Mem, we conducted grid searches. In detail, the performance of Qwen3.5 was evaluated on the even sessions of Friends dataset with new names. Table~\ref{tab:ablation_k_tkg_sess} shows that $(k_{r}, k_{s}) = (250, 40)$ is an optimal combination among all the candidates that we tried. The performance of LLM tends to improve as $k_{r}$ and $k_{s}$ increase, but it saturates beyond a certain point. Table~\ref{tab:ablation_num_seed} shows that using $n = 1$ results in the highest performance, closely followed by $n = 3$. Considering that seed nodes (i.e., the query-related candidates extracted by LLMs) act as starting point for traversing the relational memory, using $n = 1$ could be riskier than using $n = 3$; a single mismatch leaves no seed nodes and thereby retrieves no additional relational evidence in $n = 1$, whereas two other candidates still remain in $n = 3$. Therefore, we finalized $3$ as a value of $n$.

\begin{table}[htbp]
\centering
\resizebox{\columnwidth}{!}{%
\footnotesize
\begin{tabular}{cccccc}
\toprule
\diagbox{$k_{\text{s}}$}{$k_{\text{r}}$} & \textbf{50} & \textbf{150} & \textbf{250} & \textbf{350} & \textbf{450} \\

\midrule
\textbf{1}
& 71.24 \std{2.80} & 71.49 \std{1.90} & 72.00 \std{2.36} & 73.18 \std{2.36} & 72.93 \std{1.84}  \\
\textbf{5}
& 73.01 \std{1.80} & 74.03 \std{1.63} & 75.89 \std{1.52} & 75.21 \std{2.29} & 76.06 \std{2.16}  \\
\textbf{10}
& 74.62 \std{1.32} & 75.80 \std{2.55} & 76.90 \std{1.66} & 77.41 \std{1.59} & 77.66 \std{1.78}  \\
\textbf{15}
& 75.80 \std{1.69} & 77.75 \std{1.65} & 77.07 \std{1.44} & 78.60 \std{1.40} & 77.16 \std{2.75}  \\
\textbf{20}
& 75.72 \std{1.84} & 77.33 \std{1.28} & 77.16 \std{1.98} & 78.00 \std{3.46} & 78.68 \std{2.08}  \\
\textbf{25}
& 76.65 \std{2.54} & 77.75 \std{2.94} & 78.76 \std{1.06} & 77.41 \std{0.88} & 78.43 \std{0.25}  \\
\textbf{30}
& 76.82 \std{1.06} & 77.07 \std{0.96} & 78.17 \std{1.41} & 79.53 \std{0.82} & 78.43 \std{0.00}  \\
\textbf{35}
& 77.66 \std{0.88} & 78.76 \std{1.17} & 78.00 \std{0.96} & 79.95 \std{1.11} & 78.17 \std{1.11}  \\
\textbf{40}
& 78.00 \std{1.63} & 79.95 \std{1.11} & \textbf{80.37} \std{\textbf{0.82}} & 78.93 \std{1.11} & 78.93 \std{0.67}  \\
\textbf{45}
& 78.76 \std{0.53} & 77.66 \std{1.32} & 79.73 \std{0.59} & 79.53 \std{2.74} & 79.53 \std{1.61}  \\
\textbf{50}
& 79.10 \std{1.28} & 78.34 \std{3.13} & 79.02 \std{0.39} & 79.02 \std{1.40} & 78.85 \std{1.17}  \\
\bottomrule
\end{tabular}%
}
\vspace{-2mm}
\caption{Results of hyperparameter ablation on $k_{r}$ and $k_{s}$ (GroupChat-Mem).}
\vspace{-2mm}
\label{tab:ablation_k_tkg_sess}
\end{table}

\begin{table}[htbp] 
\centering
\resizebox{\columnwidth}{!}{%
\footnotesize
\begin{tabular}{cccccc}
\toprule
\textbf{$n$}  & \textbf{1} & \textbf{3} & \textbf{5} & \textbf{7} & \textbf{9} \\
\midrule
\textbf{Qwen3.5} & \textbf{76.90} \std{\textbf{0.67}} & 76.40 \std{1.78} & 75.30 \std{1.44} & 75.97 \std{0.82} & 74.96 \std{2.27}  \\
\bottomrule
\end{tabular}%
}
\vspace{-2mm}
\caption{Results of hyperparameter ablation on $n$ (GroupChat-Mem).}
\vspace{-2mm}
\label{tab:ablation_num_seed}
\end{table}

\subsection{An ablation study to evaluate each component of GroupChat-Mem}
\label{app:ablation_component}
To evaluate the contribution of each component in GroupChat-Mem, we measured performance under ablation scenarios in which a single component was removed from the full GroupChat-Mem pipeline. As shown in Table~\ref{tab:ablation_elements}, removing any module leads to a performance drop across all data types, confirming that each component contributes meaningfully to question-answering performance.

\begin{table}[htbp]
\centering
\resizebox{\columnwidth}{!}{%
\footnotesize
\begin{tabular}{lccc}
\toprule
\textbf{Method} & \textbf{Qwen3.5} & \textbf{Gemma4} & \textbf{GPT-OSS} \\
\midrule
\textbf{Ours}
& 77.83 & 77.58 & 64.81 \\
- $\mathcal{R}$
& 73.60 \textcolor{red}{\std{-4.23}} & 72.88 \textcolor{red}{\std{-4.70}} & 59.90 \textcolor{red}{\std{-4.91}} \\
- $\mathcal{S}$
& 60.74 \textcolor{red}{\std{-17.09}} & 54.57 \textcolor{red}{\std{-23.01}} & 45.64 \textcolor{red}{\std{-19.17}} \\
- Traversal through $\mathcal{R}$
& 76.73 \textcolor{red}{\std{-1.10}} & 75.04 \textcolor{red}{\std{-2.54}} & 63.83 \textcolor{red}{\std{-0.98}} \\
\bottomrule
\end{tabular}%
}
\vspace{-2mm}
\caption{Experimental results on GroupChat-Bench (Friends, New Names). $\mathcal{R}$ and $\mathcal{S}$ refer to the relational memory and raw session repository, respectively.}
\vspace{-2mm}
\label{tab:ablation_elements}
\end{table}

\section{Qualitative Analysis}
\label{app:qualitative_analysis}
\subsection{Comparing retrieved evidence from GroupChat-Mem and Baseline Methods}



\begin{figure*}[htbp]
    \centering
    \begin{center}
    \includegraphics[width=\linewidth]{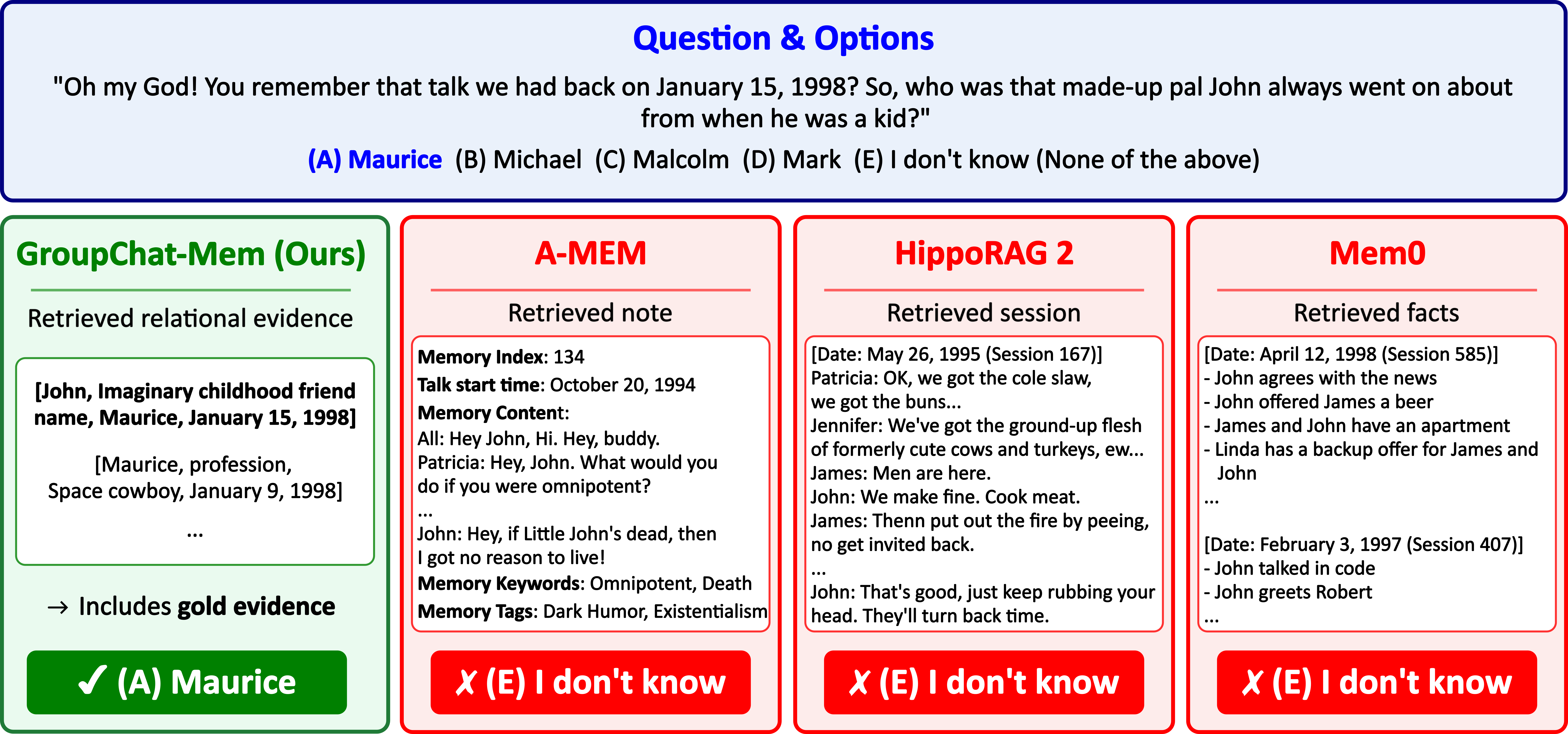}
    \end{center}
    \vspace{-2mm}
    \caption{GroupChat-Mem retrieves the dated gold memory and answers correctly, whereas all three baselines retrieve many dialogue sessions that do not contain the gold entity and therefore fail.}
    \label{fig:qualitative_analysis}
\end{figure*}

To understand the source of GroupChat-Mem's gains over the three baselines (HippoRAG 2, A-Mem, Mem0), we compare the evidence each memory system retrieves for a fixed question. We consistently observe that the evidence required to answer the question is retrieved only by GroupChat-Mem, while it is absent from the retrievals of most baselines (at least two of the three). Figure~\ref{fig:qualitative_analysis} illustrates a representative case: for the question \textbf{"Who was the made-up childhood friend John used to talk about?"}, GroupChat-Mem retrieves the gold relational evidence \textbf{[John, imaginary childhood friend name, Maurice, January 15, 1998]} in a compact, structured form and answers correctly. In contrast, A-Mem, HippoRAG 2, and Mem0 each retrieve lots of unstructured dialogue that does not contain the answer entity (Maurice), and therefore all fail. This suggests that GroupChat-Mem could retrieve more accurate and answer-bearing evidence compared to all baselines.

\subsection{Visualization of the Induced Relational Memory}


\begin{figure*}[ht]
    \centering
    \begin{center}
    \includegraphics[width=\linewidth]{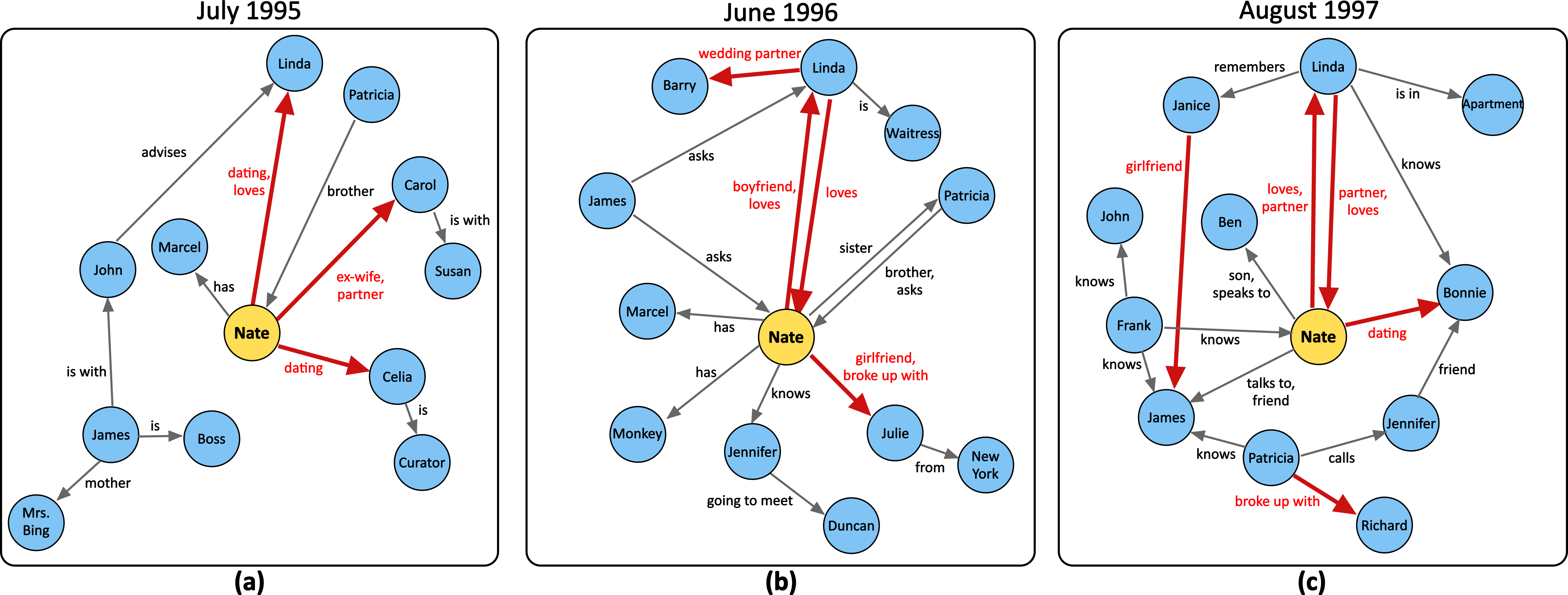}
    \end{center}
    \vspace{-4mm}
    \caption{Snapshots of the relational memory centered on one main character (\textbf{Nate}), taken across three consecutive years. Red and gray arrows represent romantic and the other relations, respectively. }
    \vspace{-3.5mm}
    \label{fig:tkg_visualization}
\end{figure*}


Figure~\ref{fig:tkg_visualization} visualizes the relational
memory~$\mathcal{R}$ induced by GroupChat-Mem, centered on one main
character from the Friends dataset under the New Names setting, with
snapshots sampled at three different time points. The visualization
illustrates two properties of $\mathcal{R}$. First, relational
evidence is consolidated into a structured form rather than dispersed
across multiple sessions. Second, the information associated with each
character is continually updated as the dialogue progresses. This
representation reduces the burden on the LLM of comprehending long,
unstructured session histories, and makes gold evidence more explicit
and easily inspectable.

\section{Annotator Instructions}
\label{app:labelingtool}
\begin{figure*}[h]
    \centering
    \begin{center}
    \includegraphics[width=0.9\linewidth]{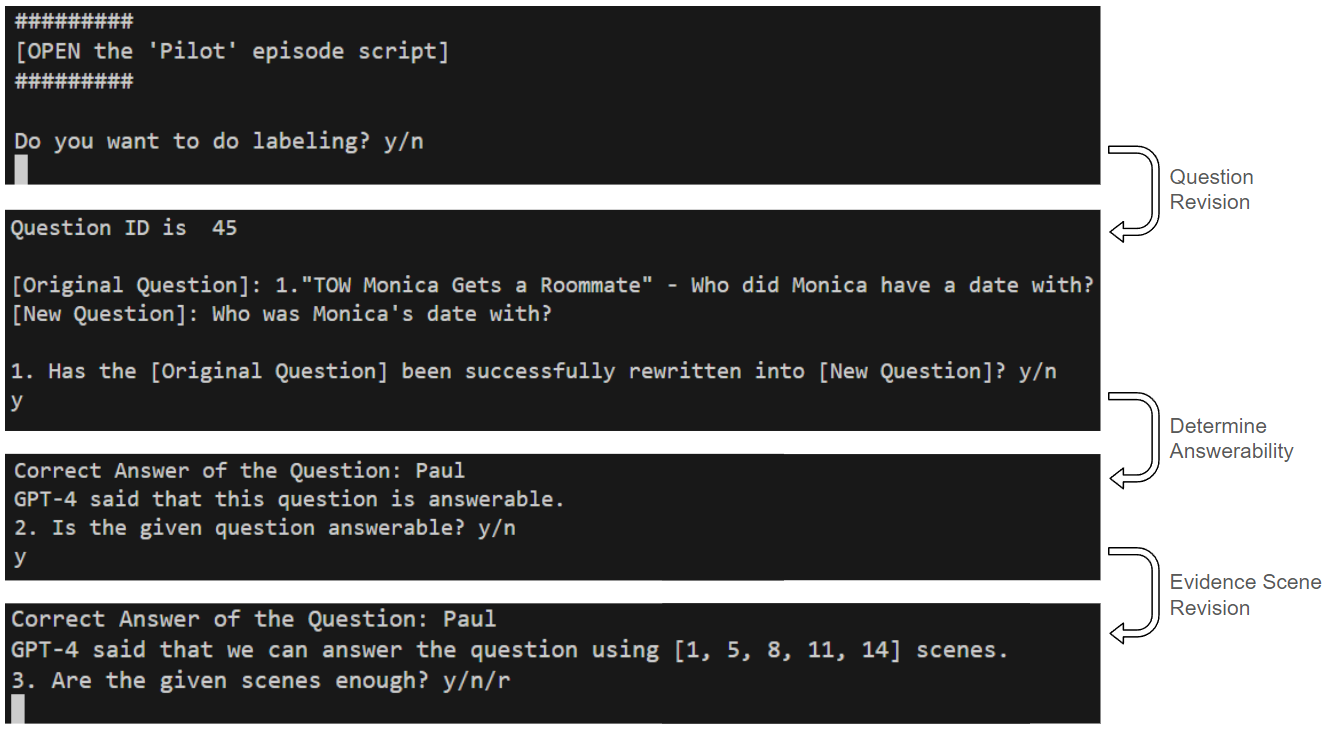}

    \end{center}
    \caption{The actual process of annotating questions from fan quizzes.}

    \label{fanquizlabeling}
\end{figure*}

\begin{figure*}[h]
    \centering
    \begin{center}
    \includegraphics[width=0.9\linewidth]{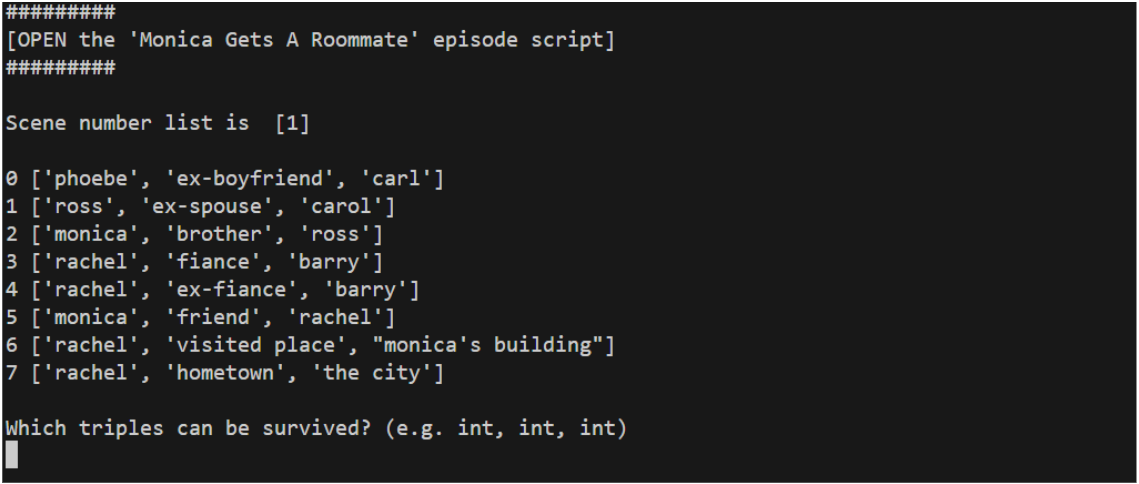}
    \end{center}
    \caption{The actual process of reviewing extracted triples.}
    \label{tkgquesitonlabeling}
\end{figure*}

Figure~\ref{fanquizlabeling} and Figure~\ref{tkgquesitonlabeling} show the screenshots of the dataset labeling process. Figure~\ref{fanquizlabeling} illustrates the annotation process for the questions based on fan quizzes, and Figure~\ref{tkgquesitonlabeling} describes the review process for selecting triples for the TKG. 

\begin{table*}[h]
\centering
\resizebox{0.85\textwidth}{!}{%
\begin{tabular}{llll}
\hline
\textbf{Question Type} & \textbf{Relation} &\textbf{Template} & \textbf{Question Example}\\ \hline
\multirow{32}{*}{\textbf{Without Time}} & alumni & Who is $\{\text{sub}\}$'s alumni? & Who is Lincoln High School's alumni? \\ \cline{2-4} 
 & boss & Who is $\{\text{sub}\}$'s boss? & Who is Chandler's boss? \\ \cline{2-4} 
 & \text{sub}ordinate & Who is $\{\text{sub}\}$'s \text{sub}ordinate? & Who is Chandler's \text{sub}ordinate? \\ \cline{2-4} 
 & client & Who is $\{\text{sub}\}$'s client? & Who is Chandler's client? \\ \cline{2-4} 
 & neighbor & Who is $\{\text{sub}\}$'s neighbor? & Who is Chandler's neighbor? \\ \cline{2-4} 
 & roommate & Who is $\{\text{sub}\}$'s roommate? & Who is Chandler's roommate? \\ \cline{2-4} 
 & ex-roommate & Who is $\{\text{sub}\}$'s ex-roommate? & Who is Chandler's ex-roommate? \\ \cline{2-4} 
 & fiance  & Who is $\{\text{sub}\}$'s fiance? & Who is Rachel's fiance? \\ \cline{2-4} 
 & fiancee  & Who is $\{\text{sub}\}$'s fiancee? & Who is Ross's fiancee? \\ \cline{2-4} 
 & ex-fiance  & Who is $\{\text{sub}\}$'s ex-fiance? & Who is Rachel's ex-fiance? \\ \cline{2-4} 
 & ex-fiancee  & Who is $\{\text{sub}\}$'s ex-fiancee? & Who is Ross's ex-fiancee? \\ \cline{2-4} 
 & pet  & Who is $\{\text{sub}\}$'s pet? & Who is Ross's pet? \\ \cline{2-4} 
 & dating with  & Who is dating $\{\text{sub}\}$? & Who is dating Ross? \\ \cline{2-4} 
 & job  & What is $\{\text{sub}\}$'s job? & What is Ross's job? \\ \cline{2-4} 
 & place of work  & Where does $\{\text{sub}\}$ work? & Where does Ross work? \\ \cline{2-4} 
 & age  & How old is $\{\text{sub}\}$? & How old is Ross? \\ \cline{2-4} 
 & major  & What is $\{\text{sub}\}$'s major? & What is Ross's major? \\ \cline{2-4} 
 & mother  & Who is $\{\text{sub}\}$'s mother? & Who is Ross's mother? \\ \cline{2-4}
 & father  & Who is $\{\text{sub}\}$'s father? & Who is Ross's father? \\ \cline{2-4}
 & place of birth  & Where was $\{\text{sub}\}$ born? & Where was Ben born? \\ \cline{2-4}
 & hometown  & Where is $\{\text{sub}\}$'s hometown? & Where is Monica's hometown? \\ \cline{2-4}
 & date of birth  & When was $\{\text{sub}\}$ born? & When was Ben born? \\ \cline{2-4}
 & husband  & Who is $\{\text{sub}\}$'s husband? & Who is Emily's husband?  \\ \cline{2-4}
 & wife  & Who is $\{\text{sub}\}$'s wife? & Who is Ross's wife? \\ \cline{2-4}
 & girlfriend  & Who is $\{\text{sub}\}$'s girlfriend? & Who is Joey's girlfriend? \\ \cline{2-4}
 & boyfriend  & Who is $\{\text{sub}\}$'s boyfriend? & Who is Monica's boyfriend? \\ \cline{2-4}
 & ex-husband  & Who is $\{\text{sub}\}$'s ex-husband? & Who is Carol's ex-husband? \\ \cline{2-4}
 & ex-wife  & Who is $\{\text{sub}\}$'s ex-wife? & Who is Ross's ex-wife? \\ \cline{2-4}
 & ex-girlfriend  & Who is $\{\text{sub}\}$'s ex-girlfriend? & Who is Ross's ex-girlfriend? \\ \cline{2-4}
 & ex-boyfriend  & Who is $\{\text{sub}\}$'s ex-boyfriend? & Who is Rachel's ex-boyfriend? \\ \cline{2-4}
 & brother  & Who is $\{\text{sub}\}$'s brother? & Who is Monica's brother? \\ \cline{2-4}
 & sister & Who is $\{\text{sub}\}$'s sister? & Who is Ross's sister? \\ \hline
\end{tabular}%
}
\caption{Templates for one-hop questions without temporal information.
\label{app:question_onehop_without_time_templates}
}
\end{table*}

\begin{table*}[h]
\centering
\resizebox{\textwidth}{!}{%
\begin{tabular}{llll}
\hline
\textbf{Question Type} & \textbf{Relation} &\textbf{Template} & \textbf{Question Example}\\ \hline
\multirow{16}{*}{\textbf{With Time}} & boss & Who was $\{\text{sub}\}$'s boss on $\{\text{time}\}$? & Who was Chandler's boss on September 26th, 1994? \\ \cline{2-4} 
 & client & Who was $\{\text{sub}\}$'s client on $\{\text{time}\}$? & Who was Chandler's client on September 26th, 1994? \\ \cline{2-4} 
 & neighbor & Who was $\{\text{sub}\}$'s neighbor on $\{\text{time}\}$? & Who was Chandler's neighbor on September 26th, 1994? \\ \cline{2-4} 
 & roommate & Who was $\{\text{sub}\}$'s roommate on $\{\text{time}\}$? & Who was Chandler's roommate on September 26th, 1994? \\ \cline{2-4} 
 & fiance  & Who was $\{\text{sub}\}$'s fiance on $\{\text{time}\}$? & Who was Rachel's fiance on September 26th, 1994? \\ \cline{2-4} 
 & fiancee  & Who was $\{\text{sub}\}$'s fiancee on $\{\text{time}\}$? & Who was Ross's fiancee on September 26th, 1994? \\ \cline{2-4} 
 & pet  & Who was $\{\text{sub}\}$'s pet on $\{\text{time}\}$? & Who was Ross's pet on September 26th, 1994? \\ \cline{2-4} 
 & dating with  & Who dated $\{\text{sub}\}$ on $\{\text{time}\}$? & Who dated Ross on September 26th, 1994? \\ \cline{2-4} 
 & job  & What was $\{\text{sub}\}$'s job on $\{\text{time}\}$? & What was Monica's job on September 26th, 1994? \\ \cline{2-4} 
 & place of work  & Where did $\{\text{sub}\}$ work on $\{\text{time}\}$? & Where did Monica work on September 26th, 1994? \\ \cline{2-4} 
 & age  & How old was $\{\text{sub}\}$ on $\{\text{time}\}$? & How old was Monica on September 26th, 1994? \\ \cline{2-4} 
 & major  & What was $\{\text{sub}\}$'s major on $\{\text{time}\}$? & What was Ross's major on September 26th, 1994? \\ \cline{2-4} 
 & husband  & Who was $\{\text{sub}\}$'s husband on $\{\text{time}\}$? & Who was Emily's husband on September 26th, 1994? \\ \cline{2-4}
 & wife  & Who was $\{\text{sub}\}$'s wife on $\{\text{time}\}$? & Who was Ross's wife on September 26th, 1994? \\ \cline{2-4}
 & girlfriend  & Who was $\{\text{sub}\}$'s girlfriend on $\{\text{time}\}$? & Who was Ross's girlfriend on September 26th, 1994? \\ \cline{2-4}
 & boyfriend  & Who was $\{\text{sub}\}$'s boyfriend on $\{\text{time}\}$? & Who was Rachel's boyfriend on September 26th, 1994? \\ \hline
\end{tabular}%
}
\caption{Templates for one-hop questions with temporal information.
\label{app:question_onehop_time_templates}
}
\end{table*}

\begin{table*}[h]
\centering
\resizebox{\textwidth}{!}{%
\begin{tabular}{llll}
\hline
\textbf{First Relation} & \textbf{Second Relation} & \textbf{Template} & \textbf{Question Example} \\ \hline
\multirow{17}{*}{\begin{tabular}[c]{@{}l@{}}roommate, wife, husband, \\ girlfriend, boyfriend, client, \\ neighbor, boss, subordinate, \\ fiance, fiancee\end{tabular}} & \begin{tabular}[c]{@{}l@{}}roommate, wife, husband, pet,\\ girlfriend, boyfriend, client, neighbor,\\ boss, subordinate, fiance, fiancee\end{tabular} & \begin{tabular}[c]{@{}l@{}}\{sub1\} had a \{First Relation\} on \{time1\}. \\ Who was the \{Second Relation\} of the \\ \{First Relation\} on \{time2\}?\end{tabular} & \begin{tabular}[c]{@{}l@{}}Monica had a roommate on September 26th, 1994.\\ Who was the boyfriend of the roommate \\ on October 5th, 1996?\end{tabular} \\ \cline{2-4} 
 & dating with & \begin{tabular}[c]{@{}l@{}}\{sub1\} had a \{First Relation\} on \{time1\}. \\ Who dated the \{First Relation\} on \{time2\}?\end{tabular} & \begin{tabular}[c]{@{}l@{}}Monica had a roommate on September 26th, 1994. \\ Who dated the roommate on October 5th, 1996?\end{tabular} \\ \cline{2-4} 
 & job, major, age & \begin{tabular}[c]{@{}l@{}}\{sub1\} had a \{First Relation\} on \{time1\}. \\ What was the \{Second Relation\} of the\\ \{First Relation\} on \{time2\}?\end{tabular} & \begin{tabular}[c]{@{}l@{}}Monica had a roommate on September 26th, 1994.\\ What was the job of the roommate \\ on October 5th, 1996?\end{tabular} \\ \cline{2-4} 
 & \begin{tabular}[c]{@{}l@{}}mother, father, son, daughter, \\ sister, brother\end{tabular} & \begin{tabular}[c]{@{}l@{}}\{sub1\} had a \{First Relation\} on \{time1\}. \\ Who is the \{Second Relation\} of the\\ \{First Relation\}?\end{tabular} & \begin{tabular}[c]{@{}l@{}}Monica had a roommate on September 26th, 1994.\\ Who is the mother of the roommate?\end{tabular} \\ \cline{2-4} 
 & date of birth, place of birth, & \begin{tabular}[c]{@{}l@{}}\{sub1\} had a \{First Relation\} on \{time1\}. \\ When (Where) was the \{First Relation\} born?\end{tabular} & \begin{tabular}[c]{@{}l@{}}Monica had a roommate on September 26th 1994. \\ When was the roommate born?\end{tabular} \\ \cline{2-4} 
 & place of work & \begin{tabular}[c]{@{}l@{}}\{sub1\} had a \{First Relation\} on \{time1\}. \\ Where did the \{First Relation\} work \\ on \{time2\}?\end{tabular} & \begin{tabular}[c]{@{}l@{}}Monica had a roommate on September 26th, 1994. \\ Where did the roommate work \\ on October 5th, 1996?\end{tabular} \\ \cline{2-4} 
 & hometown & \begin{tabular}[c]{@{}l@{}}\{sub1\} had a \{First Relation\} on \{time1\}.\\ Where is the hometown of the \\ \{First Relation\}?\end{tabular} & \begin{tabular}[c]{@{}l@{}}Monica had a roommate on September 26th, 1994.\\ Where is the hometown of the roommate?\end{tabular} \\ \hline
dating with & \begin{tabular}[c]{@{}l@{}}roommate, wife, husband, \\ girlfriend, boyfriend, client, neighbor,\\ boss, subordinate, fiance, fiancee\end{tabular} & \begin{tabular}[c]{@{}l@{}}\{sub1\} dated a person on \{time1\}. \\ Who was the \{Second Relation\} of the \\ person on \{time2\}?\end{tabular} & \begin{tabular}[c]{@{}l@{}}Monica dated a person on September 26th, 1994.\\ Who was the boss of the person \\ on October 5th, 1996?\end{tabular} \\ \hline
\multirow{13}{*}{\begin{tabular}[c]{@{}l@{}}mother, father, son, \\ daughter, sister, brother\end{tabular}} & \begin{tabular}[c]{@{}l@{}}roommate, wife, husband, \\ girlfriend, boyfriend, client, neighbor, \\ boss, subordinate, fiance, fiancee\end{tabular} & \begin{tabular}[c]{@{}l@{}}Who was the \{Second Relation\} of \{sub1\}'s \\ \{First Relation\} on \{time2\}?\end{tabular} & \begin{tabular}[c]{@{}l@{}}Who was the roommate of Ross's \\ sister on September 26th, 1994?\end{tabular} \\ \cline{2-4} 
  & dating with & \begin{tabular}[c]{@{}l@{}}Who dated \{sub1\}'s \{First Relation\} \\ on \{time2\}?\end{tabular} & Who dated Ben's father on September 26th, 1994? \\ \cline{2-4} 
 & job, age, major & \begin{tabular}[c]{@{}l@{}}What was the \{Second Relation\} of \\ \{sub1\}'s \{First Relation\} on \{time2\}?\end{tabular} & What was the job of Ben's father on September 26th, 1994? \\\cline{2-4} 
  & \begin{tabular}[c]{@{}l@{}}mother, father, son, daughter, \\ sister, brother\end{tabular} & \begin{tabular}[c]{@{}l@{}}Who is the \{Second Relation\} of \\ \{sub1\}'s \{First Relation\}?\end{tabular} & Who is the mother of Ross's son? \\ \cline{2-4} 
 & date of birth, place of birth & \begin{tabular}[c]{@{}l@{}}When (Where) was \{sub1\}'s \{First Relation\} \\ born?\end{tabular} & When was Monica's brother born? \\ \cline{2-4} 
 & place of work & \begin{tabular}[c]{@{}l@{}}Where did \{sub1\}'s \{First Relation\} \\ work on \{time2\}?\end{tabular} & \begin{tabular}[c]{@{}l@{}}Where did Monica's brother work \\ on October 5th, 1996?\end{tabular} \\ \cline{2-4} 
 & hometown & \begin{tabular}[c]{@{}l@{}}Where is the hometown of \\ \{sub1\}'s \{First Relation\}?\end{tabular} & Where is the hometown of Ross's son? \\ 
 \hline
\end{tabular}%
}
\caption{Templates for two-hop questions.
\label{app:question_twohop_templates}
}
\end{table*}

\begin{table*}[h]
\centering
\resizebox{\textwidth}{!}{%
\begin{tabular}{lll}
\hline
\textbf{Original Question} & \textbf{Character} & \textbf{Style Transferred Question} \\ \hline
\multirow{9}{*}{By the way, how did Rachel buy her new boots?} & Default & \begin{tabular}[c]{@{}l@{}}Hey, any idea what Rachel used to snag those stylish \\ new boots of hers?\end{tabular} \\ \cline{2-3} 
 & Monica & \begin{tabular}[c]{@{}l@{}}Hey, do you know what Rachel used to snag those super \\ cute new boots she's been rocking?\end{tabular} \\ \cline{2-3} 
 & Chandler & \begin{tabular}[c]{@{}l@{}}So, could we BE any more curious about how Rachel \\ snagged those new boots?\end{tabular} \\ \cline{2-3} 
 & Joey & \begin{tabular}[c]{@{}l@{}}Hey, how did Rachel manage to snag those killer \\ boots, huh?\end{tabular} \\ \cline{2-3} 
 & Phoebe & \begin{tabular}[c]{@{}l@{}}Oh my gosh! Do you have any idea how Rachel snagged\\  those super cute new boots?\end{tabular} \\  \hline
\multirow{9}{*}{By the way, who dated Monica on September 22, 1994?} & Default & \begin{tabular}[c]{@{}l@{}}So, who was Monica's date on the night of \\ September 22, 1994?\end{tabular} \\ \cline{2-3} 
 & Chandler & \begin{tabular}[c]{@{}l@{}}Oh, could you BE any more specific about who was \\ going out with Monica on September 22, 1994?\end{tabular} \\ \cline{2-3} 
 & Joey & \begin{tabular}[c]{@{}l@{}}Hey, just outta curiosity, who was goin' out with \\ Monica on September 22, 1994?\end{tabular} \\ \cline{2-3} 
 & Phoebe & \begin{tabular}[c]{@{}l@{}}Oh my gosh, so like, who was Monica's date on that \\ super specific day, September 22, 1994?\end{tabular} \\ \cline{2-3} 
 & Rachel & \begin{tabular}[c]{@{}l@{}}Oh my god, so like, who was going out with Monica \\ on September 22, 1994?\end{tabular} \\ \hline
\multirow{14}{*}{\begin{tabular}[c]{@{}l@{}}By the way, Rachel had a roommate on October 28, 1994. \\ Who dated the roommate in September 1994?\end{tabular}} & Default & \begin{tabular}[c]{@{}l@{}}Oh. My. God. Remember when Rachel had a roommate \\ back on October 28, 1994? So, who was going out with\\  that roommate by September 1994?\end{tabular} \\ \cline{2-3} 
 & Monica & \begin{tabular}[c]{@{}l@{}}Hey, just out of curiosity, do you know who was going \\ out with Rachel's roommate from back in September 1994? \\ I remember she got that roommate around October 28, 1994.\end{tabular} \\ \cline{2-3} 
 & Chandler & \begin{tabular}[c]{@{}l@{}}So, just for a little stroll down memory lane, Rachel \\ was bunking with someone on October 28, 1994. \\ Any wild guesses on who was dating this mystery \\ co-habitant by September 1994?\end{tabular} \\ \cline{2-3} 
 & Joey & \begin{tabular}[c]{@{}l@{}}Hey, so you know how Rachel was living with someone\\  back on October 28, 1994, right? So I'm just wonderin' \\ here, who was going out with this roommate of hers in \\ September 1994?\end{tabular} \\ \cline{2-3} 
 & Phoebe & \begin{tabular}[c]{@{}l@{}}By the way, Rachel had a roommate on October 28, 1994. \\ Who dated the roommate in September 1994?\end{tabular} \\ \hline
\end{tabular}%
}
\caption{Examples of the results of character style transfer.
\label{app:question_styletransfer}
}
\end{table*}

\section{Prompts}
\label{app:prompts}

Figures ~\ref{fig:openie}--\ref{fig:qa_open_ended} present the prompts used in the GroupChat-Mem experiments, Figure~\ref{fig:hippo_openie}--\ref{fig:hippo_qa} present those used for HippoRAG~2, Figure~\ref{fig:mem0_fact_extraction}--\ref{fig:mem0_qa} present those used for Mem0, and Figure~\ref{fig:amem_note_construction}--\ref{fig:amem_qa} present those used for A-MEM.

For brevity, some prompts are truncated, with omitted content denoted as ``\ldots''; the full prompts are available at \url{https://github.com/jwyang21/groupchat_bench/}.

\clearpage
\onecolumn
\begingroup
    \centering
    \tcbinputlisting{promptbox, listing file={prompt/bridge/preprocess/openie-triple-extraction.txt}}
    \captionof{figure}{Prompt used for OpenIE.}
    \label{fig:openie}
\endgroup
\twocolumn

\clearpage
\onecolumn
\begingroup
    \centering
    \tcbinputlisting{promptbox, listing file={prompt/bridge/seednode_identification/stage1_batch.txt}}
    \captionof{figure}{Prompt used to identify candidates of query-related TKG nodes.}
    \label{fig:seednode_stage1}
\endgroup
\twocolumn

\clearpage
\onecolumn
\begingroup
    \centering
    \tcbinputlisting{promptbox, listing file={prompt/bridge/seednode_identification/stage2_finalize.txt}}
    \captionof{figure}{Prompt used to finalize query-related TKG nodes.}
    \label{fig:seednode_stage2}
\endgroup
\twocolumn

\clearpage
\onecolumn
\begingroup
    \centering
    \tcbinputlisting{promptbox, listing file={prompt/bridge/iteration_lookup/bridge.txt}}
    \captionof{figure}{Prompt used to identify which entity to look up for TKG traversal.}
    \label{fig:iteration_lookup}
\endgroup
\twocolumn

\clearpage
\onecolumn
\begingroup
    \centering
    \tcbinputlisting{promptbox, listing file={prompt/bridge/final_qa/multi-structured/enhanced.txt}}
    \captionof{figure}{Prompt used for multi-structured question-answering.}
    \label{fig:qa_multi_structured}
\endgroup
\twocolumn

\clearpage
\onecolumn
\begingroup
    \centering
    \tcbinputlisting{promptbox, listing file={prompt/bridge/final_qa/multi-unstructured/enhanced.txt}}
    \captionof{figure}{Prompt used for multi-unstructured question answering.}
    \label{fig:qa_multi_unstructured}
\endgroup
\twocolumn

\clearpage
\onecolumn
\begingroup
    \centering
    \tcbinputlisting{promptbox, listing file={prompt/bridge/final_qa/open-ended/enhanced.txt}}
    \captionof{figure}{Prompt used for open-ended question answering.}
    \label{fig:qa_open_ended}
\endgroup
\twocolumn

\clearpage
\onecolumn
\begingroup
    \centering
    \tcbinputlisting{promptbox, listing file={prompt/hipporag2/stage1_openie.txt}}
    \captionof{figure}{Prompt used for triple extraction in HippoRAG 2.}
    \label{fig:hippo_openie}
\endgroup
\twocolumn

\clearpage
\onecolumn
\begingroup
    \centering
    \tcbinputlisting{promptbox, listing file={prompt/hipporag2/stage2_filter-before-qa.txt}}
    \captionof{figure}{Prompt used to select query-related triples in HippoRAG 2.}
    \label{fig:hippo_filter}
\endgroup
\twocolumn

\clearpage
\onecolumn
\begingroup
    \centering
    \tcbinputlisting{promptbox, listing file={prompt/hipporag2/stage3_qa.txt}}
    \captionof{figure}{Prompt used for question-answering in HippoRAG 2.}
    \label{fig:hippo_qa}
\endgroup
\twocolumn

\clearpage
\onecolumn
\begingroup
    \centering
    \tcbinputlisting{promptbox, listing file={prompt/mem0/stage1_fact-extraction.txt}}
    \captionof{figure}{Prompt used for extracting salient facts in Mem0.}
    \label{fig:mem0_fact_extraction}
\endgroup
\twocolumn

\clearpage
\onecolumn
\begingroup
    \centering
    \tcbinputlisting{promptbox, listing file={prompt/mem0/stage2_update-memory.txt}}
    \captionof{figure}{Prompt used for memory update in Mem0.}
    \label{fig:mem0_memory_update}
\endgroup
\twocolumn

\clearpage
\onecolumn
\begingroup
    \centering
    \tcbinputlisting{promptbox, listing file={prompt/mem0/qa.txt}}
    \captionof{figure}{Prompt used for question-answering in Mem0.}
    \label{fig:mem0_qa}
\endgroup
\twocolumn

\clearpage
\onecolumn
\begingroup
    \centering
    \tcbinputlisting{promptbox, listing file={prompt/amem/stage1_note-construction.txt}}
    \captionof{figure}{Prompt used for note construction in A-MEM.}
    \label{fig:amem_note_construction}
\endgroup
\twocolumn

\clearpage
\onecolumn
\begingroup
    \centering
    \tcbinputlisting{promptbox, listing file={prompt/amem/stage2_evolution.txt}}
    \captionof{figure}{Prompt used to link and update memory notes in A-MEM.}
    \label{fig:amem_memory_update}
\endgroup
\twocolumn

\clearpage
\onecolumn
\begingroup
    \centering
    \tcbinputlisting{promptbox, listing file={prompt/amem/stage3_linking-notes.txt}}
    \captionof{figure}{Prompt used to generate retrieval keywords from the question in A-MEM (following the official implementation).}
    \label{fig:amem_link_update}
\endgroup
\twocolumn

\clearpage
\onecolumn
\begingroup
    \centering
    \tcbinputlisting{promptbox, listing file={prompt/amem/qa.txt}}
    \captionof{figure}{Prompt used for question-answering in  A-MEM.}
    \label{fig:amem_qa}
\endgroup
\twocolumn

\end{document}